\documentclass[onecolumn]{robbyant}

\metadata[Website]{\url{https://technology.robbyant.com/lingbot-map}}
\metadata[Github]{\url{https://github.com/robbyant/lingbot-map}}

\newcommand\methodname{LingBot-Map}

\newcommand{\method}{\texttt{\methodname}\xspace}

\usepackage{overpic}
\usepackage{fontawesome}
\usepackage{bbding}
\usepackage{wasysym}
\usepackage{utfsym}
\usepackage{arydshln}
\usepackage{colortbl}
\usepackage{gensymb}
\usepackage{appendix}
\usepackage{makecell}
\usepackage{tikz}
\usepackage{float}
\usetikzlibrary{tikzmark, calc}

\algtext*{EndWhile}
\algtext*{EndFor}
\algtext*{EndIf}
\algtext*{EndFunction}
\algtext*{EndProcedure}

\definecolor{tabfirst}{rgb}{1, 0.7, 0.7}
\definecolor{tabsecond}{rgb}{1, 0.85, 0.7}
\definecolor{tabthird}{rgb}{1, 1, 0.7}
\definecolor{tabgray}{rgb}{0.9, 0.9, 0.9}
\definecolor{MyDarkRed}{rgb}{0.66, 0.16, 0.16}
\definecolor{MyDarkBlue}{rgb}{0.16, 0.16, 0.66}

\newcommand{\secref}[1]{Sec.~\ref{#1}}

\makeatletter
\DeclareRobustCommand\onedot{\futurelet\@let@token\@onedot}
\def\@onedot{\ifx\@let@token.\else.\null\fi\xspace}
\def\eg{\emph{e.g}\onedot} 
\def\ie{\emph{i.e}\onedot}

\makeatother

\title{LingBot-Map: Geometric Context Transformer  \\[6pt]  for Streaming 3D Reconstruction}

\author{
\begin{center}
    Lin-Zhuo Chen$^{*}$ \quad
    Jian Gao$^{*}$ \quad
    Shangzhan Zhang$^{*}$ \quad
    Yihang Chen \quad
    Ka Leong Cheng \quad
    Yipengjing Sun \quad \\ [6pt]
    Liangxiao Hu \quad
    Nan Xue \quad
    Xing Zhu \quad
    Yujun Shen \quad
    Yao Yao$^{\dagger}$ \quad
    Yinghao Xu$^{\dagger \ddagger}$
    \\[12pt]
    $^{*}$Equal Contribution \qquad
    $^{\dagger}$Corresponding Author \qquad
    $^{\ddagger}$Project Lead
\end{center}
}

\hypersetup{
  pdftitle={LingBot-Map: Geometric Context Transformer for Streaming 3D Reconstruction},
  pdfauthor={Lin-Zhuo Chen, Jian Gao, Shangzhan Zhang, Yihang Chen, Ka Leong Cheng, Yipengjing Sun, Liangxiao Hu, Nan Xue, Xing Zhu, Yujun Shen, Yao Yao, Yinghao Xu},
  pdflang={en-US}
}

\begin{document}

\abstract{%
Streaming 3D reconstruction aims to recover 3D information, such as camera poses and point clouds, from a video stream, which necessitates geometric accuracy, temporal consistency, and computational efficiency.
Motivated by the principles of Simultaneous Localization and Mapping (SLAM), we introduce \method, a feed-forward 3D foundation model for reconstructing scenes from streaming data, built upon a \textit{geometric context transformer} (GCT) architecture.
A defining aspect of \method lies in its carefully designed attention mechanism, which integrates an anchor context, a pose-reference window, and a trajectory memory to address coordinate grounding, dense geometric cues, and long-range drift reduction, respectively.
This design keeps the streaming state compact while retaining rich geometric context, enabling stable efficient inference at around 20 FPS on $518 \times 378$ resolution inputs over long sequences exceeding 10,000 frames.
Extensive evaluations across a variety of benchmarks demonstrate that our approach achieves superior performance compared to both existing streaming and iterative optimization-based approaches.
}

\maketitle

\begin{figure}[h]
  \centering
  \begin{overpic}[width=1.\columnwidth, trim={0 0 0 0}, clip]{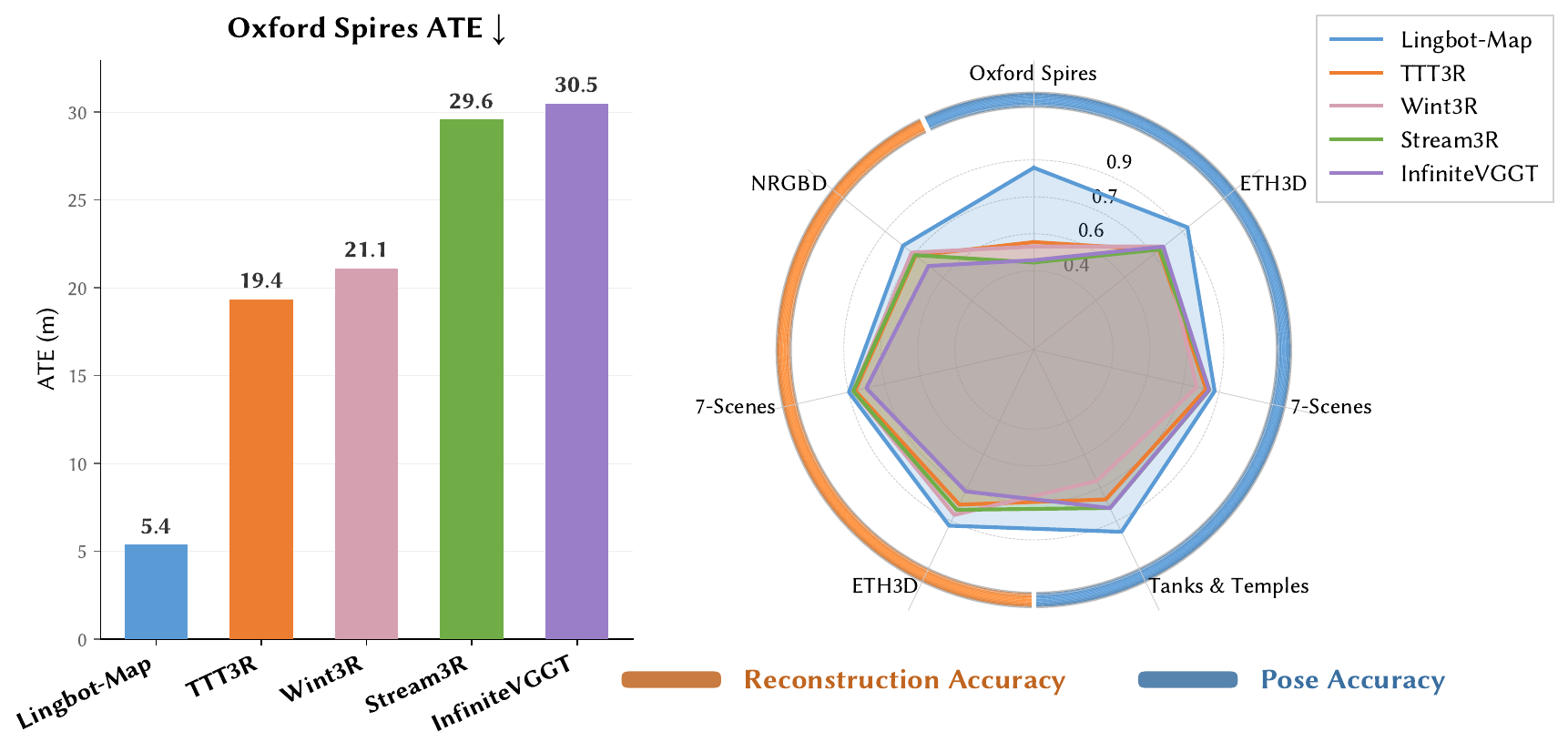}
	\end{overpic}
  \caption{\textbf{Comparison of \method with State-of-the-Art streaming reconstruction methods.}}

  \label{fig:teasor}
\end{figure}

\justifying

\section{Introduction}
\label{sec:intro}

\begin{figure}[p]
  \centering
  \begin{overpic}[width=1.\columnwidth, trim={30 140 40 40}, clip]{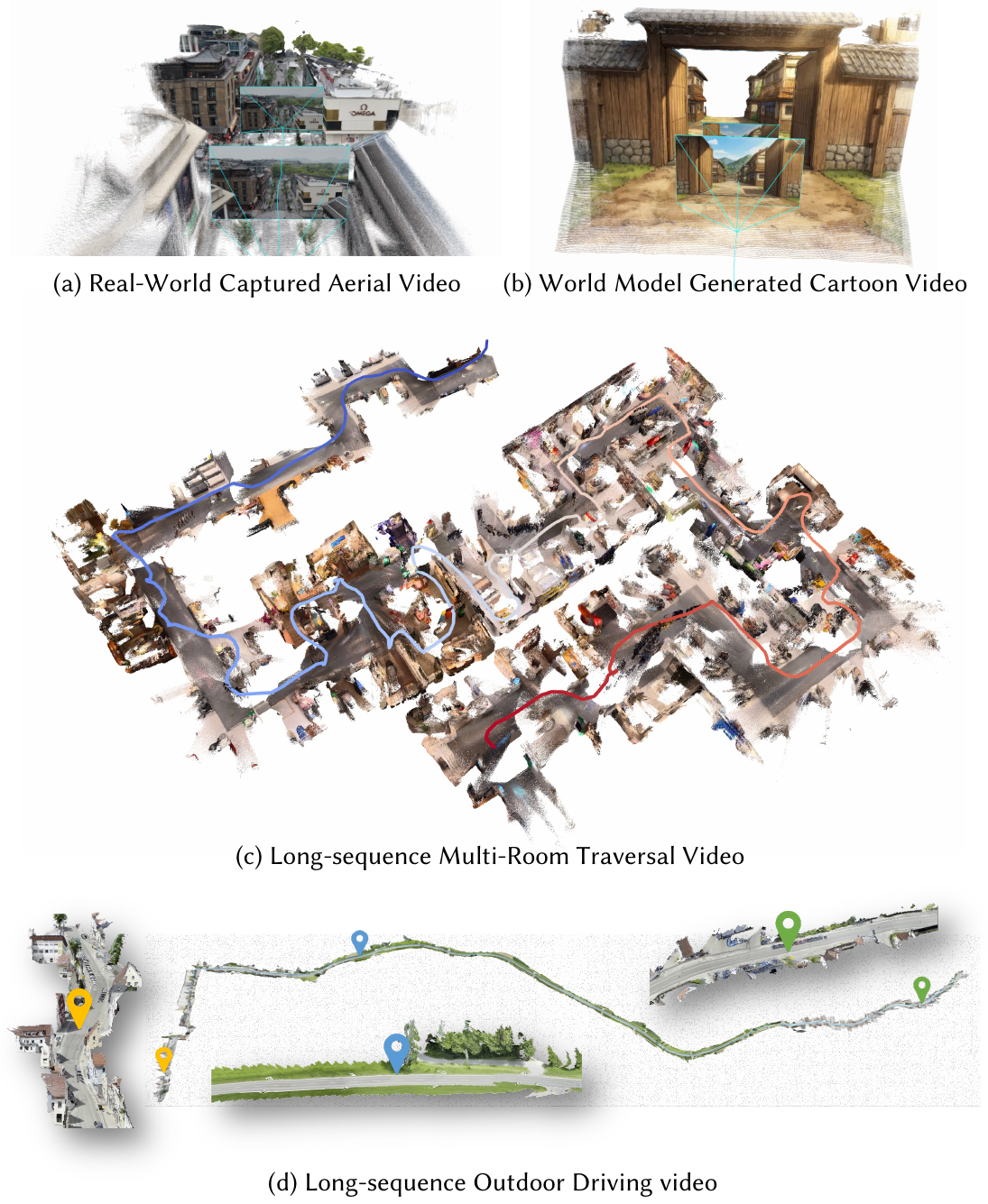}
	\end{overpic}
  \caption{\textbf{Visualization results of \method in VO mode on varying scenes:} Given a continuous video stream, \method performs \textit{efficient} \textbf{streaming} 3D reconstruction with accurate camera pose estimation and high-quality point cloud reconstruction. Examples include Real-World Captured Aerial Video (\emph{top-left}), World Model Generated Cartoon Video (\emph{top-right}), Long-sequence Multi-Room Traversal Video (\emph{middle}), and Long-sequence Outdoor Driving video (\emph{bottom}).}
  \label{fig:visualization}
\end{figure}

We perceive the world through a continuous stream of visual input, yet our spatial memory is not a faithful recording of every moment: it is sparse, structured, and efficient.
Rather than retaining every observation, human spatial cognition selectively preserves only the most essential cues, enabling coherent navigation through large-scale environments over extended periods.
Can we build machines that operate similarly, performing streaming 3D reconstruction from continuous visual input, selectively and efficiently?

In recent years, 3D foundation models have advanced rapidly, with methods such as VGGT~\cite{wang2025vggt} and Depth Anything 3~\cite{depthanything3} directly predicting camera poses, depth maps, and dense point maps from multiple images in a single feed-forward pass.
However, these successes are largely confined to offline settings, where the full image set is available and processed globally.
Recent efforts~\cite{cut3r,streamVGGT,stream3r2025,li2025wint3r} have begun to adapt these capabilities to streaming reconstruction, but existing approaches still exhibit limited robustness to long sequences and complex scenes.
The core difficulty is selective context management: balancing rich geometric context for long-term consistency with a compact state for efficient inference.

Existing methods adopt different strategies to manage streaming context, each involving distinct trade-offs.
CUT3R~\cite{cut3r} maintains a persistent recurrent-style state, but its aggressive compression can lead to state forgetting and weak retention of essential geometric priors.
In contrast, StreamVGGT~\cite{streamVGGT} and Stream3R \cite{stream3r2025} adopt causal attention and caching, yet retain near-complete history without explicit selection, mixing useful geometry with redundancy and causing memory and computation to grow rapidly.
A third line of work, such as VGGT-SLAM~\cite{vggtslam} and MASt3R-SLAM~\cite{mast3rslam}, integrates learned 3D models with classical SLAM backends. While these methods incorporate structured context management through keyframe selection and pose-graph maintenance, such selection relies on hand-crafted heuristics rather than learned priors, and iterative optimization further limits real-time applicability.
These observations point to a key design principle: the streaming state should selectively retain \emph{what} matters most, not merely how much, and this selection should be grounded in geometric priors yet learned end-to-end from data.

To this end, we introduce \method, a streaming foundation model built around Geometric Context Attention (GCA) that realizes this principle within a unified attention framework.
The design draws on a key insight from classical SLAM systems: robust real-time reconstruction requires maintaining distinct types of spatial context---a reference frame for coordinate grounding, a local window for dense local geometry estimation, and a global map for long-range drift reduction.
Accordingly, GCA explicitly maintains three complementary types of context: an \emph{anchor context} for coordinate and scale grounding, a local \emph{pose-reference window} that retains dense visual features from recent frames for accurate local geometry estimation, and a \emph{trajectory memory} that compresses the full observation history into compact per-frame tokens for global consistency.
While the context structure is motivated by classical reconstruction principles, GCA replaces hand-crafted optimization with end-to-end learned attention that adaptively weights, encodes, and compresses information within each context type.
This structured yet learned representation ensures stable and efficient inference even over arbitrarily long sequences, with nearly constant memory and computation per frame, as context beyond the local window is compressed into compact per-frame tokens.

To scale up training effectively, we employ a progressive training strategy combined with context parallelism~\cite{jacobs2023deepspeed}, along with a relative loss formulation that facilitates stable optimization on long sequences.
This allows \method to learn efficiently from diverse, large-scale 3D datasets.
We evaluate \method on comprehensive benchmarks, including Oxford Spires, 7-Scenes, Tanks and Temples, ETH3D, and show consistent improvements over existing streaming methods in both camera pose estimation and dense 3D reconstruction quality.

Our contributions are summarized as follows:
\begin{itemize}
    \item We introduce \method, a streaming 3D foundation model built around Geometric Context Attention (GCA), which maintains three complementary context types -- anchor, pose-reference window, and trajectory memory -- for efficient and consistent long-sequence streaming inference.
    \item We propose an efficient training recipe based on progressive training and context parallelism with a relative loss formulation for stable long-sequence optimization.
    \item We demonstrate that \method achieves state-of-the-art performance on multiple benchmarks (Oxford Spires, Tanks and Temples, ETH3D, and 7-Scenes), significantly outperforming existing streaming approaches in reconstruction quality and inference speed.
\end{itemize}

\section{Related Work}
\label{sec:related}

\noindent\textbf{Traditional 3D Reconstruction.}
Traditional 3D reconstruction methods mainly include Structure-from-Motion (SfM)~\cite{snavely2006photo, schonberger2016structure, pan2024global}, Simultaneous Localization and Mapping (SLAM)~\cite{mur2015orb,mur2017orb,campos2021orb}, and Multi-View Stereo (MVS)~\cite{furukawa2009accurate,schonberger2016pixelwise,yao2018mvsnet}. SfM and SLAM recover camera poses and scene geometry from multi-view observations, where SfM typically operates offline on unordered image collections, while SLAM processes video streams online. These systems are usually complex and highly modular, typically centered around optimization-based bundle adjustment for camera pose estimation. In contrast, MVS focuses on dense reconstruction given known camera poses. Over the past decade, many works have explored replacing individual components of these pipelines with deep learning modules, particularly for feature extraction~\cite{detone2018superpoint} and matching~\cite{sarlin2020superglue,sun2021loftr}. More recently, several approaches attempt to implement SfM, SLAM, or MVS in an end-to-end manner, such as VGGTSfM~\cite{wang2024vggsfm}, DROID-SLAM~\cite{teed2021droid}, and MVSNet~\cite{yao2018mvsnet,yao2019recurrent}.

\noindent\textbf{3D Foundation Model.}
DUSt3R~\cite{dust3r} represents a paradigm shift in feed-forward 3D reconstruction.
Given a collection of unposed images, DUSt3R directly regresses a dense 3D reconstruction of the scene without explicit geometric modeling.
However, DUSt3R~\cite{dust3r} only supports two-view input and requires aligning all results via optimization for additional views.
To support more than two views and improve reconstruction quality, VGGT~\cite{wang2025vggt} uses an advanced transformer architecture that includes cross-view attention layers, achieving state-of-the-art performance on standard benchmarks.
Crucially, VGGT demonstrates that leveraging large-scale data together with powerful model architectures can significantly improve reconstruction quality.
Building on this foundation, numerous subsequent works have advanced feed-forward reconstruction across various dimensions, including improving reconstruction accuracy~\cite{wang2025vggt, zhang2025flare, wang2025pi, depthanything3}, enhancing computational efficiency~\cite{yang2025fast3r, liu2025slam3r}, handling dynamic scenes~\cite{zhang2024monst3r, sucar2025dynamic, chen2025easi3r, feng2025st4rtrack, xiao2025spatialtracker, lu2025align3r, jin2025stereo4d, wang2025spatialvid}, enabling novel view synthesis~\cite{lin2025longsplat, ye2024no, jiang2025anysplat, chen2025long3r, smart2024splatt3r, xu2024grm, hong2024lrm, tang2024lgm, xu2024dmv3d, li2024instant3d, wang2024pflrm}, and incorporating multi-modal inputs~\cite{lu2025matrix3d, jang2025pow3r, keetha2025mapanything}.
However, these methods are primarily designed for offline processing and do not address the unique challenges of streaming 3D reconstruction, such as maintaining long-term consistency and managing computational resources over extended sequences.

\noindent\textbf{Streaming 3D Reconstruction.}
Driven by the need for online applications, streaming 3D reconstruction can be broadly categorized into hybrid SLAM-based approaches and end-to-end feed-forward methods.
Hybrid methods typically integrate 3D foundation models with traditional SLAM pipelines~\cite{vggtslam, mast3rslam, deng2025vggt, xie2026scal3r}, aiming to leverage the strengths of both paradigms.
However, these approaches often rely on hand-crafted components and careful parameter tuning to achieve optimal performance, lacking the benefits of a fully end-to-end learning framework.
In contrast, recent feed-forward streaming methods~\cite{stream3r2025, li2025wint3r, streamVGGT, wang20253d, cut3r} extend the offline paradigm to streaming settings by employing Recurrent Neural Network (RNN)-based architectures or by combining caching mechanisms with causal attention.
Specifically, CUT3R~\cite{cut3r} maintains a persistent state that is updated recurrently via RNN architectures.
To mitigate state forgetting, TTT3R~\cite{chen2025ttt3r} adopts a test-time training strategy.
Meanwhile, StreamVGGT~\cite{streamVGGT}, Stream3R~\cite{stream3r2025}, and Wint3R~\cite{li2025wint3r} adapt the more advanced VGGT architecture using causal attention and caching strategies.
Despite these advances, existing streaming methods often struggle to maintain performance over long input sequences and in complex environments.
Common failure modes include significant trajectory drift, degraded reconstruction accuracy, and prohibitive growth in memory and computational requirements.
We attribute these limitations to the absence of a principled mechanism for effectively retaining essential geometric context during the streaming process.
Concurrently, LoGeR~\cite{zhang2026loger} and Scal3R~\cite{xie2026scal3r} explore scaling 3D reconstruction to long sequences via chunk-based processing with hybrid memory mechanisms.
LoGeR combines sliding window attention for local alignment with test-time training (TTT) for global consistency, while Scal3R extends the TTT paradigm with chunking and visual place recognition for large-scale scenes.
However, both methods operate in an offline setting and rely on test-time parameter updates, which introduces additional computational overhead.
In contrast, our \method is a purely feed-forward streaming model that requires no test-time training or post-optimization, achieving real-time inference through a compact geometric context design.

\section{Method}
\label{sec:method}

In~\cref{sec:overview}, we present an overview of our method and the problem formulation.
We then introduce Geometric Context Attention (GCA) (\cref{sec:attention}), a mechanism specifically designed for streaming 3D reconstruction. \cref{sec:network} describes the network architecture and the training procedure.
Finally,~\cref{sec:inference} presents the inference pipeline for efficient inference.

\subsection{Overview}
\label{sec:overview}
\begin{figure}[t]
  \centering
  \begin{overpic}[width=1\columnwidth]{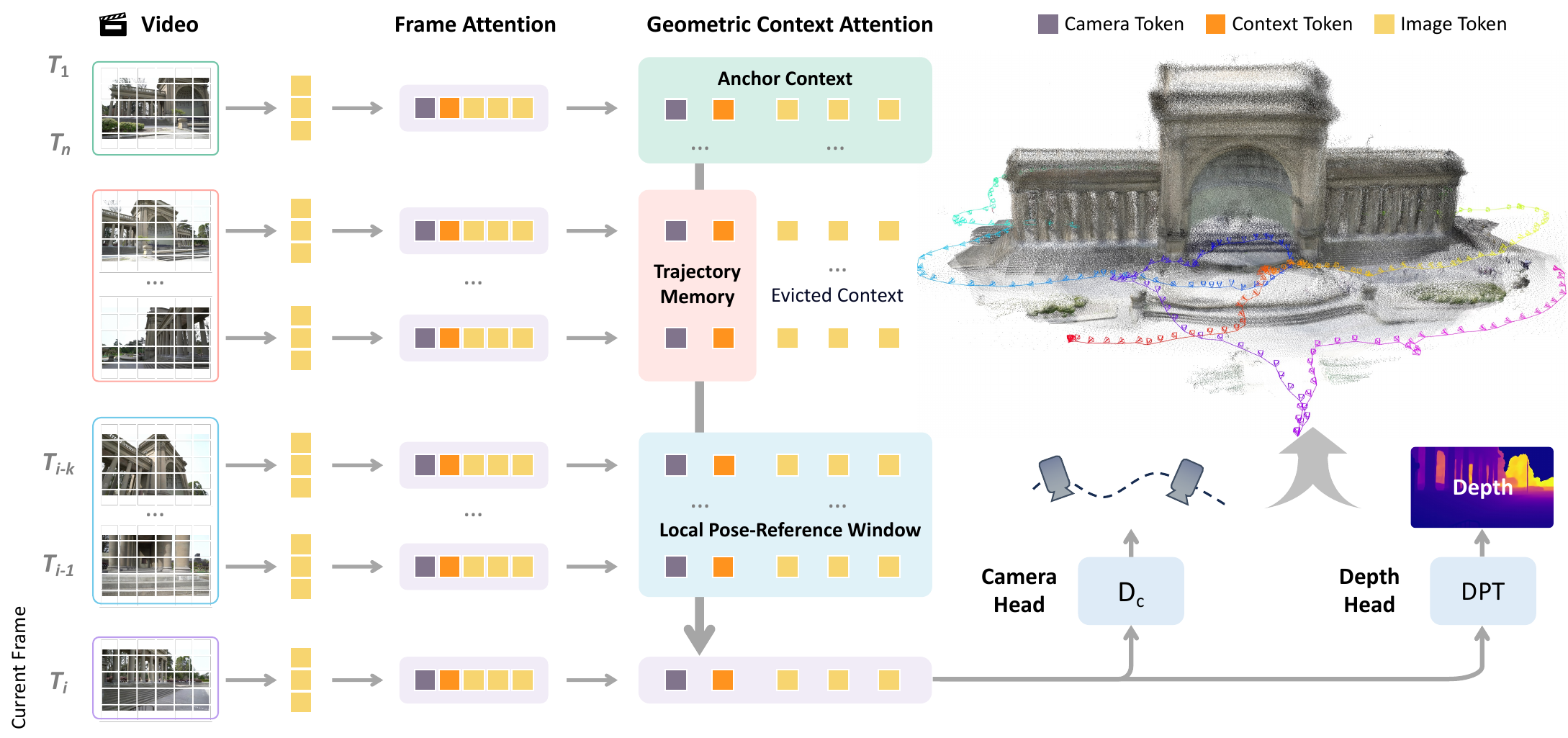}
    \end{overpic}
  \caption{\textbf{Pipeline of the proposed \method.} The framework processes the current view $T_i$ relative to an initialization set $[T_1, T_n)$. A DINO backbone extracts image features, which are then refined through alternating layers of Frame Attention and GCA. Within the GCA module, the input view aggregates information from the \textit{Anchor Context}, \textit{Local Pose-Reference Window }$[T_{i-k}, T_i)$, and \textit{Trajectory Memory Context}. Finally, task-specific heads predict camera pose and depth maps, enabling robust, memory-efficient streaming 3D reconstruction for long-range sequences.}
  \label{fig:GCT}
\end{figure}

Given a continuous stream of images $\mathcal{I} = \{I_1, I_2, \ldots\}$, \method processes each new frame $I_t$ upon arrival and estimates its camera pose $\hat{P}_t$ and depth map $\hat{D}_t$ using only the current and previously observed frames $\{I_1, \ldots, I_t\}$, without access to future observations.
Building on the architecture of recent feed-forward 3D models~\cite{wang2025vggt}, we design a streaming variant where each frame is encoded by a ViT backbone and processed through alternating layers of frame-wise attention and Geometric Context Attention (GCA), before task-specific heads predict the camera pose and depth map (see~\cref{fig:GCT}).
The key to enabling efficient streaming inference is GCA, which maintains three complementary geometric contexts: anchor, pose-reference window, and trajectory memory, thereby balancing long-term consistency with compact state representation.
We detail GCA in~\cref{sec:attention}, the overall architecture and training strategy in~\cref{sec:network}, and the inference pipeline in~\cref{sec:inference}.

\subsection{Geometric Context Attention}
\label{sec:attention}
The central challenge of streaming 3D reconstruction is managing geometric context: the model must retain sufficient long-range context to ensure global consistency, yet keep its streaming state compact enough for efficient inference.
Classical SLAM and SfM systems provide structural insights into this trade-off by decomposing the streaming state into three distinct types of spatial context, each serving a complementary role: a reference frame for coordinate and scale grounding, a local window of recent observations for dense geometry estimation, and a global map for reducing accumulated drift.
Drawing on this principle, GCA decomposes the streaming context into three complementary learned attention mechanisms, replacing hand-crafted optimization with end-to-end differentiable attention: \emph{anchor context}, a local \emph{pose-reference window}, and \emph{trajectory memory}.
We describe each below.

\noindent\textbf{Anchor Context.}
Monocular reconstruction is inherently scale-ambiguous, so a consistent coordinate system and absolute scale must be established before streaming begins.
Offline methods such as DUSt3R~\cite{dust3r} and VGGT~\cite{wang2025vggt} resolve this by normalizing with respect to the \emph{global} point cloud, but this requires access to all frames and is therefore incompatible with causal streaming inference.
Instead, we designate the first $n$ images ($n \ll N$) as anchor frames and use them to fix the scale.
We apply full attention among these frames. Every frame carries a learnable anchor token, but anchor frames and streaming frames use \emph{different} learned values for this token, so the network can identify and distinguish anchor frames from subsequent streaming frames.
After initialization, the anchor tokens and image tokens of the anchor frames are retained in the attention context, and all subsequent frames attend to them as fixed references.
We refer to all information carried by these initialization frames collectively as the \emph{anchor context}.
During training, we normalize all ground-truth annotations to a canonical scale derived from the anchor frames:
we compute $s = \frac{1}{|\bar{\mathcal{X}}^{\text{anchor}}|} \sum_{\mathbf{x}\in\bar{\mathcal{X}}^{\text{anchor}}} \lVert\mathbf{x}\rVert_2$ as the mean distance of the ground-truth point cloud $\bar{\mathcal{X}}^{\text{anchor}}$ from the coordinate origin, and divide all ground-truth depths and camera translations by $s$.

\noindent\textbf{Local Pose-Reference Window.}
Accurately registering each new frame requires dense visual overlap with nearby observations, context that the distant anchor frames alone cannot provide.
To address this, we maintain a sliding window of the $k$ most recent frames during inference, retaining their \emph{full} image tokens.
This dense local context provides essential relative pose cues from immediate visual connections, enabling the network to accurately register new frames into the global trajectory.
To further encourage geometric consistency within the local window, we apply a relative pose loss between frames in this window, as detailed in~\cref{sec:network}.

\noindent\textbf{Trajectory Memory.}
The anchor context and local window together provide a fixed global reference and dense recent observations, but without any record of intermediate frames, pose errors accumulate unchecked over long sequences, causing the estimated trajectory to drift.
To mitigate this, we retain a compact trajectory context that summarizes the full observation history.
Specifically, for frames that fall outside both the anchor set and the active sliding window, we retain only the camera, anchor, and register tokens (i.e.,~6 context tokens per frame) while discarding the memory-intensive image tokens ($M$ tokens per frame).
Additionally, we incorporate video temporal positional encodings~\cite{wan2025} into the retained tokens to impose temporal ordering on the global trajectory.
By maintaining this lightweight yet temporally ordered record of all past observations, the trajectory memory provides long-range cues that help correct accumulated drift and ensure global consistency.

\begin{figure}[t]
  \centering
  \begin{overpic}[width=0.9\columnwidth, trim={30 10 30 0}, clip]{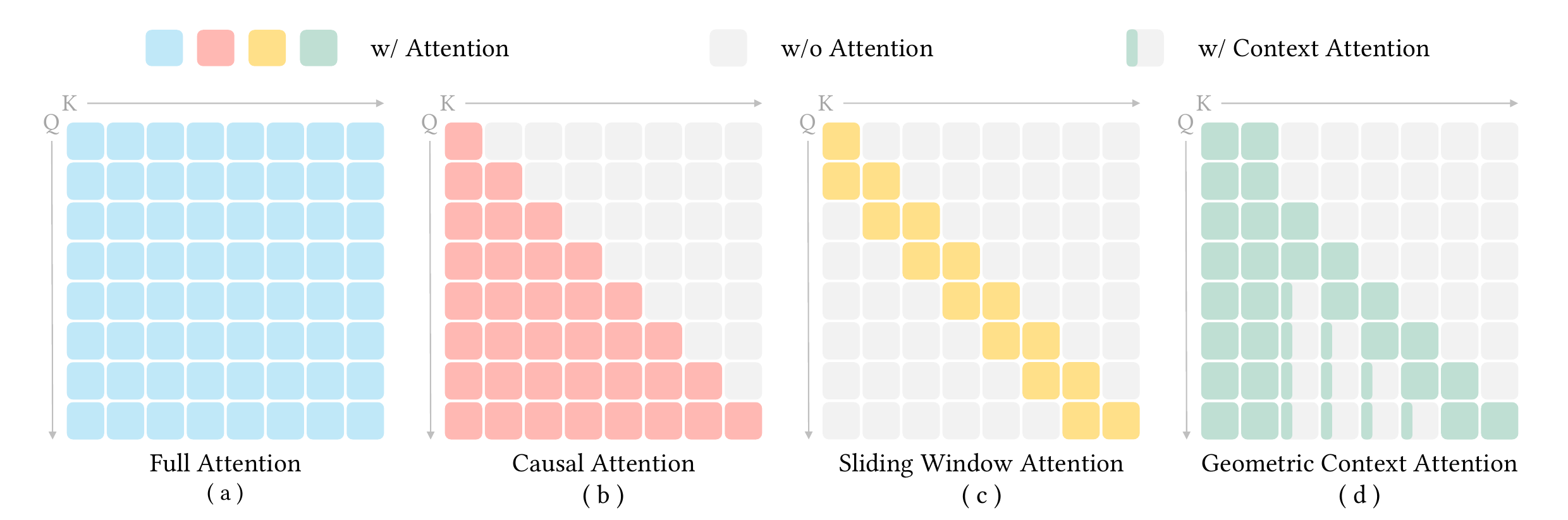}
	\end{overpic}
        \caption{\textbf{Comparison of attention masks.} Each square block represents one frame's tokens, consisting of a small leading segment of context tokens (camera, anchor, and register tokens) and a larger segment of image tokens. (a)~Full attention attends to all frames. (b)~Causal attention enables streaming but grows linearly with sequence length. (c)~Sliding-window attention bounds cost but loses long-range context. (d)~GCA partitions the streaming context into \textit{anchors} ($n{=}2$), a \textit{local window} ($k{=}2$), and \textit{trajectory memory}, retaining rich long-range context while keeping cost nearly constant as the sequence length increases.}

  \label{fig:attention_map}
\vspace{-8pt}
\end{figure}

\noindent\textbf{Attention Mask Design.}
\cref{fig:attention_map} compares different attention patterns for streaming inference.
Global attention~(a) attends to all frames but cannot operate in a streaming fashion.
Causal attention~(b) enables streaming but causes memory and computation to grow linearly with sequence length.
Sliding window attention~(c) bounds compute but sacrifices long-term context.
Our GCA~(d) combines the anchor context, trajectory memory, and local window into a structured attention mask that retains long-range consistency with bounded per-frame cost.

\noindent\textbf{Complexity Analysis.}
For a $T$-frame sequence, the per-frame attention context in GCA comprises $n$ anchor frames with full tokens ($n \cdot (M + 6)$), $k$ window frames with full tokens ($k \cdot (M + 6)$), and $(T - n - k)$ trajectory frames with compact tokens ($6$ each).
Since $n$ and $k$ are fixed constants, the total context simplifies to $(n + k) \cdot M + 6T$, where the first term is constant and the second grows at a rate of just 6 tokens per frame.
In contrast, causal attention retains $T \cdot (M + 6) = MT + 6T$ tokens, sharing the same $6T$ term but incurring an additional $MT$ that grows with the full token count.
Since each new frame adds $(M{+}6)$ tokens under causal attention but only $6$ under GCA, the per-frame growth rate is reduced by roughly $70{\times}$ for typical values ($M {\approx} 500$).
Concretely, with $n {=} 3$, $k {=} 16$, and $T {=} 10{,}000$, causal attention accumulates ${\sim}5{\times}10^{6}$ tokens, while GCA retains only ${\sim}7{\times}10^{4}$, yielding nearly constant memory and computation per frame.

\subsection{Geometric Context Transformer Framework}
\label{sec:network}
\noindent\textbf{Architecture.}
The overall architecture is illustrated in~\cref{fig:GCT}.
Each input image is first encoded by a DINOv2-Large Vision Transformer (ViT) backbone~\cite{oquab2023dinov2} to produce $M$ image tokens per frame.
These image tokens are augmented with a camera token $\mathbf{c} \in \mathbb{R}^{C}$, four register tokens $\mathbf{r}_j \in \mathbb{R}^{C}$ ($j = 1, \ldots, 4$), and a learnable anchor token $\mathbf{a} \in \mathbb{R}^{C}$.
Following VGGT~\cite{wang2025vggt}, the register tokens act as shared scratch tokens that participate in attention but are not decoded by the prediction heads.
The augmented tokens are then processed through multiple alternating layers of Frame Attention and GCA.
Frame Attention operates independently within each frame, enabling per-frame feature refinement, while GCA operates across frames according to the structured attention mask described in~\cref{sec:attention}, enabling cross-frame geometric reasoning.
Finally, a camera head takes the camera token to predict the absolute camera pose $\hat{P}_t$, and a depth head takes the image tokens to predict the corresponding depth map $\hat{D}_t$.

\noindent\textbf{Loss Function.}
We train \method using a composite loss function consisting of depth, absolute pose, and relative pose terms:
\begin{equation}
\mathcal{L}
= \lambda_{\text{depth}} \, \mathcal{L}_{\text{depth}}
+ \lambda_{\text{abs-pose}} \, \mathcal{L}_{\text{abs-pose}}
+ \lambda_{\text{rel-pose}} \, \mathcal{L}_{\text{rel-pose}}.
\end{equation}

\noindent The depth loss ($\mathcal{L}_{\text{depth}}$) and absolute pose loss ($\mathcal{L}_{\text{abs-pose}}$)
follow the definitions in VGGT~\cite{wang2025vggt}: 
$
  \mathcal{L}_{\text{depth}}
= \sum_{i=1}^{N} \Big\|
\Sigma_i^{D} \odot (\hat{D}_i - D_i) \Big\|
  + \Big\| \Sigma_i^{D} \odot (\nabla \hat{D}_i - \nabla D_i) \Big\|
  - \alpha \log \Sigma_i^{D},
  \mathcal{L}_{\text{abs-pose}}
= \sum_{i=1}^{N} \left\lVert \hat{\mathbf{P}}_{i} - \mathbf{P}_{i} \right\rVert_{\epsilon}.
$
Here, $\Sigma_i^{D}$ represents the predicted uncertainty map, and $\odot$ denotes element-wise multiplication.
Unlike VGGT~\cite{wang2025vggt}, we supervise the network using camera-to-world transformations rather than world-to-camera ones. In the world-to-camera parameterization, rotation and translation are inherently coupled, making translation estimation highly sensitive to rotation errors, particularly in long sequences.
Inspired by $\pi^3$~\cite{wang2025pi}, we incorporate a relative pose loss over all frame pairs within the sliding window:
\begin{equation}
  \mathcal{L}_{\text{rel-pose}}
= \frac{1}{k (k - 1)}
  \sum_{\substack{i \neq j \\ i,j \in \{1,\ldots,k\}}} \bigl(
    \mathcal{L}_{\text{rot}}(i,j)
    + \lambda_{\text{trans}} \mathcal{L}_{\text{trans}}(i,j)
  \bigr),
\end{equation}
where $\mathcal{L}_{\text{rot}}(i,j)$ and $\mathcal{L}_{\text{trans}}(i,j)$ denote the geodesic rotation error and $\ell_1$ translation error of the relative pose between frames $i$ and $j$, respectively.
Because the window comprises only already-observed frames, this loss is inherently causal and encourages local trajectory consistency.

\noindent\textbf{Progressive View Training.}
Training directly on long sequences is challenging: early-stage pose errors propagate along the trajectory and destabilize the loss landscape, leading to slow or divergent optimization.
To address this, we adopt a progressive training strategy that starts with short subsequences and gradually increases the number of views over training.
This curriculum enables the network to first acquire reliable local geometry estimation from short clips before learning to maintain global consistency across progressively longer trajectories.

\noindent\textbf{Context Parallel.}
As the number of training views grows, GPU memory becomes the primary bottleneck due to the quadratic cost of cross-frame attention.
To address this, we employ the Ulysses~\cite{jacobs2023deepspeed} context-parallelism strategy, which distributes different views across multiple GPUs to enable parallel attention computation via efficient all-to-all collective communication.

\subsection{Inference Pipeline}
\label{sec:inference}

\noindent\textbf{Inference design.}
Like autoregressive LLMs, our causal model caches the key-value (KV) states of previously processed frames to avoid redundant recomputation. However, with naive causal attention, the KV cache scales linearly with the number of frames, increasing memory consumption and per-frame latency.
GCA addresses this by keeping the per-frame context compact (see~\cref{sec:attention}), but the sliding-window and trajectory-eviction logic still requires frequent cache updates (appending new entries and discarding old ones), which incur overhead due to repeated memory reallocation under a standard contiguous layout.
We eliminate this overhead with a paged KV-cache layout~\cite{kwon2023efficient}, in which updates affect only newly appended tokens rather than the entire cached sequence.

We implement the runtime on FlashInfer~\cite{ye2025flashinfer}, which provides native support for paged KV-cache management, as well as optimized attention kernels for paged and sparse KV layouts.
In the $518 \times 378$ setting with video sequences up to 1000 frames and a sliding window of 64 frames, our FlashInfer-based implementation achieves $\sim$20 FPS, compared to $\sim$10.5 FPS for an otherwise identical PyTorch baseline with contiguous KV-cache updates.
When an input exceeds the maximum sequence length used during training, we retain one keyframe every $m$ frames in the KV cache to support robust long-sequence inference.

\section{Training \& Inference}
\label{sec:training}

Training a streaming 3D reconstruction model end-to-end on long sequences is
challenging: pose errors in early frames propagate along the trajectory and
destabilize the loss landscape, making direct optimization on hundreds of views
impractical.
To address this, we adopt a two-stage training curriculum.
The first stage trains an offline base model on short, diverse multi-view data
to establish robust geometric priors~(\cref{sec:base_training}).
The second stage introduces our proposed Geometric Context Attention (GCA) and
progressively scales to long sequences, transferring the base model's
geometric foundations to the streaming setting~(\cref{sec:streaming_training}).
We describe the training data curation in~\cref{sec:training_data} and
the inference pipeline in~\cref{sec:inference_mode}.

\subsection{Base Model Training}
\label{sec:base_training}

\paragraph{Model Initialization.}
We initialize the ViT backbone from DINOv2-Large~\cite{oquab2023dinov2}
with a patch size of 14 pixels, followed by 24 alternating blocks of
frame attention and cross-frame attention, following the architecture
of VGGT~\cite{wang2025vggt}.
At this stage, we use standard global attention rather than GCA:
since the training data includes both unordered multi-view collections and
temporally ordered video sequences, global attention imposes no temporal
structure and can fully exploit both data types.
The number of input views per training sample is randomly sampled between
2 and 24, matching the diverse scale of available datasets.

\paragraph{Optimization.}
We use AdamW with a base learning rate of $2 \times 10^{-4}$ and weight
decay of 0.05.
The learning rate follows a linear warmup from $10^{-8}$ to the base rate
over the first 5\% of training, followed by cosine annealing back to
$10^{-8}$ over the remaining 95\%.
Training runs for 160K iterations.

\paragraph{Data Augmentation.}
Images are resized to a maximum dimension of 518 pixels.
To improve robustness to appearance variation across datasets with diverse
capture conditions, we apply aggressive photometric augmentation:
random color jittering (brightness, contrast, saturation $\pm 0.5$;
hue $\pm 0.1$) with probability 0.9, random Gaussian blur, random
grayscale conversion with probability 0.05, and random spatial rescaling
in $[0.8\times, 1.2\times]$ with aspect ratios sampled from $[0.33, 1.0]$.
Additionally, we apply \emph{co-jittering}---applying an identical color
transform to all frames within a scene---with probability 0.3, and
independent per-frame transforms otherwise.
This encourages the model to rely on geometric cues rather than
appearance shortcuts when frames share similar photometric
characteristics, while the independent transforms build robustness to
inter-frame appearance variation.

\paragraph{Distributed Training.}
This stage takes approximately two weeks on 64 GPUs, or about 21,500 GPU
hours. We use fully sharded data parallelism (FSDP) with gradient
checkpointing and bfloat16 mixed precision to manage memory consumption.

\subsection{Streaming Model Training}
\label{sec:streaming_training}

\paragraph{Initialization from Base Model.}
We initialize the streaming model from the pretrained base model weights
and replace global attention with GCA.
Since the query, key, and value projections in GCA share the same
parameterization as global attention, the pretrained weights transfer
directly, providing a strong initialization that accelerates convergence.

\paragraph{Optimization and Progressive Curriculum.}
We train for 160K iterations with AdamW at a base learning rate of
$5 \times 10^{-5}$, using the same warmup and cosine annealing schedule
as the first stage.
To stabilize training on progressively longer sequences, we adopt a
view curriculum: the number of training views increases linearly from 24
to 320 over the course of training.
The starting count of 24 views matches the maximum used in the base
stage, and the upper limit of 320 is set by the GPU memory budget under
context parallelism.
Similarly, the local pose reference-window size $k$ of GCA is randomly sampled from 16 to 64
during training, exposing the model to varying receptive fields and
improving robustness at inference time when different window sizes may
be used.

\paragraph{Context Parallelism.}
As the number of views grows, GPU memory becomes the primary bottleneck
due to the quadratic cost of cross-frame attention.
We employ the Ulysses~\cite{jacobs2023deepspeed} context-parallelism
strategy with a parallelism dimension of 16, distributing different views
across GPUs and computing attention via all-to-all collective
communication.
Our implementation builds on TorchTitan~\cite{liang2025torchtitan} and
Magi Attention.
This stage takes approximately one week on 64 GPUs.

\subsection{Training Data}
\label{sec:training_data}

\paragraph{Dataset Composition.}
The training corpus comprises the 31 dataset entries listed in
\cref{tab:dataset_composition}, spanning indoor, outdoor, object-centric,
synthetic, and real-world scenarios.
The full list, together with data format, scene type, and per-stage
sampling ratios, is given in~\cref{tab:dataset_composition}.
The entries use three source formats: unordered multi-view collections,
temporally ordered video sequences, and mesh-based datasets. The
distinction between unordered and temporally continuous inputs drives
the sampling strategies used in each training stage.

\paragraph{Stage 1: Diverse Short-Sequence Data.}
The first stage aims to build general geometric priors from a broad
distribution of scenes.
Stage~1 uses 26 of these entries, with the sampling ratios listed in
\cref{tab:dataset_composition}.
The multi-view collections include
BlendedMVS~\cite{yao2020blendedmvs},
HyperSim~\cite{roberts2021hypersim},
MegaDepth~\cite{li2018megadepth},
MVS-Synth~\cite{huang2018deepmvs},
GTA-SFM~\cite{wang2020flow},
and CO3D~\cite{reizenstein2021common}.
The mesh-based entries are Objaverse~\cite{deitke2023objaverse} and
Texverse~\cite{zhang2025texverse}.
The video datasets include
Unreal4K~\cite{tosi2021smd},
WildRGBD~\cite{xia2024rgbd},
TartanAir~\cite{wang2020tartanair},
TartanAirV2~\cite{wang2020tartanair},
TartanGround~\cite{patel2025tartanground},
PointOdyssey~\cite{zheng2023pointodyssey},
VirtualKITTI~\cite{cabon2020virtual},
Kubric~\cite{greff2022kubric},
DL3DV~\cite{ling2024dl3dv},
SceneRGBD~\cite{McCormac:etal:ICCV2017},
Mapfree~\cite{arnold2022map},
Aria Synthetic Environments~\cite{pan2023aria},
ADT~\cite{pan2023aria},
ScanNet~\cite{dai2017scannet},
ScanNet++~\cite{yeshwanth2023scannet++},
MatrixCity~\cite{li2023matrixcity},
MidAir~\cite{Fonder2019MidAir},
and our internal game dataset.
Each iteration samples 2 to 24 frames per scene, with a dynamic batch
sampler that packs at most 48 images per GPU.
Frame selection relies on a \emph{nearby sampler}: a reference frame is
chosen at random, and the remaining frames are drawn from a spatial
window around it, without enforcing any temporal order.
This unordered sampling is well-suited to the mixed-modality data in
this stage, where many datasets have no natural frame ordering.

\paragraph{Stage 2: Long-Trajectory Video Data.}
The second stage shifts the distribution toward long, temporally coherent
sequences needed for streaming reconstruction.
We significantly increase the sampling weights for datasets with
extended trajectories and multi-scene coverage, including
TartanAir~\cite{wang2020tartanair},
TartanAirV2,
TartanGround~\cite{patel2025tartanground},
MidAir~\cite{Fonder2019MidAir},
MatrixCity~\cite{li2023matrixcity},
Waymo~\cite{sun2020scalability},
VirtualKITTI~\cite{cabon2020virtual},
KITTI-360~\cite{liao2022kitti},
ScanNet++~\cite{yeshwanth2023scannet++},
ScanNet~\cite{dai2017scannet},
and our internal game dataset.
Stage~2 also introduces the mesh-based Gibson~\cite{xia2018gibson},
Matterport3D~\cite{chang2017matterport3d}, and
HM3D~\cite{ramakrishnan2021hm3d}, while down-weighting or dropping
multi-view-only datasets that lack temporal structure
(see~\cref{tab:dataset_composition} for exact ratios).

\paragraph{Foldback Video Sampler.}
To produce temporally coherent training subsequences from long videos,
we replace the spatial nearby sampler with a \emph{foldback video
sampler}.
The sampler starts at a random frame and advances with a random stride.
Upon reaching a sequence boundary, it reverses direction and draws a new
stride (distinct from the previous one) to avoid degenerate oscillation.
This mechanism yields subsequences with naturally varying frame rates
and no forward-time bias, providing the model with diverse temporal
contexts during training.

\begin{table}[t]
\centering
\caption{Dataset composition, data format, scene type, and sampling proportions across the two training stages. Blank entries indicate that the dataset was not used in that stage.}
\label{tab:dataset_composition}
\begin{tabular}{llc | cc}
\toprule
Dataset & Format & Scene Type & Stage 1 Ratio (\%) & Stage 2 Ratio (\%) \\
\midrule
BlendedMVS~\cite{yao2020blendedmvs}     & Multi-view & Single & 1.1  & 1.9  \\
HyperSim~\cite{roberts2021hypersim}       & Multi-view & Single & 5.8  & 5.7  \\
MegaDepth~\cite{li2018megadepth}      & Multi-view & Multi  & 3.9  & -    \\
Unreal4K~\cite{tosi2021smd}       & Video      & Single & 1.0  & 0.9  \\
WildRGBD~\cite{xia2024rgbd}       & Video      & Single & 5.8  & 1.9  \\
TartanAir~\cite{wang2020tartanair}      & Video      & Multi  & 3.9  & 7.6  \\
TartanAirV2~\cite{wang2020tartanair}    & Video      & Multi  & 5.8  & 10.8 \\
TartanGround~\cite{patel2025tartanground} & Video     & Multi  & 5.8  & 10.8 \\
Waymo~\cite{sun2020scalability}        & Video      & Multi  & -    & 0.9  \\
PointOdyssey~\cite{zheng2023pointodyssey}   & Video      & Single & 0.3  & 0.9  \\
VirtualKITTI~\cite{cabon2020virtual}  & Video      & Multi  & 0.8  & 1.9  \\
Kubric~\cite{greff2022kubric}         & Video      & Single & 0.3  & 0.9  \\
DL3DV~\cite{ling2024dl3dv}          & Video      & Multi  & 11.0 & 5.7  \\
MVS-Synth~\cite{huang2018deepmvs}      & Multi-view & Single & 1.7  & -    \\
GTA-SFM~\cite{wang2020flow}      & Multi-view & Single & 1.7  & -    \\
CO3D~\cite{reizenstein2021common}          & Multi-view & Single & 5.5  & -    \\
SceneRGBD~\cite{McCormac:etal:ICCV2017}     & Video      & Single & 7.3  & 5.7  \\
Mapfree~\cite{arnold2022map}       & Video      & Multi  & 3.9  & 1.5  \\
Aria Synthetic~\cite{pan2023aria} & Video      & Single & 7.3  & 5.7  \\
ADT~\cite{pan2023aria}            & Video      & Single & 1.0  & -    \\
Objaverse~\cite{deitke2023objaverse}      & Mesh & Single & 2.7  & -    \\
Texverse~\cite{zhang2025texverse}       & Mesh & Single & 2.7  & -    \\
ScanNet++~\cite{yeshwanth2023scannet++}      & Video      & Single & 3.9  & 2.8  \\
ScanNet~\cite{dai2017scannet}        & Video      & Single & 1.9  & 2.8  \\
MatrixCity~\cite{li2023matrixcity}     & Video      & Multi  & 1.7  & 7.6  \\
MidAir~\cite{Fonder2019MidAir}         & Video      & Multi  & 2.9  & 5.7  \\
KITTI-360~\cite{liao2022kitti}      & Video      & Multi  & -    & 3.8  \\
Internal Game  & Video      & Multi  & 10.6 & 10.8 \\
Gibson~\cite{xia2018gibson}  & Mesh      & Multi  & - & 1.3 \\
Matterport3D~\cite{chang2017matterport3d}  & Mesh      & Multi  & - & 1.2 \\
HM3D~\cite{ramakrishnan2021hm3d}  & Mesh      & Multi  & - & 1.2 \\
\bottomrule
\end{tabular}%
\end{table}

\subsection{Inference Modes}
\label{sec:inference_mode}

\method supports two inference modes, Direct Output and Visual
Odometry (VO), that share a common keyframe selection mechanism.

\paragraph{Keyframe Selection.}
When the input sequence exceeds the maximum training length, we retain
one keyframe every $m$ frames in the KV cache to control context growth.
This fixed-stride keyframe rule is shared by both inference modes.

\paragraph{Direct Output Mode.}
Direct mode is the default inference setting.
The model processes frames causally through GCA, with the full
three-level context (anchor, trajectory memory, and local window)
accumulating continuously without reset.
Each frame directly outputs an absolute camera pose and a dense
depth map.
In this mode, prediction errors accumulate solely from the model's
frame-by-frame inference, without introducing any additional error
from external alignment steps.
Although the model is trained on sequences of up to 320 views, we
empirically find that the Direct mode with keyframe selection remains
stable for approximately $10\times$ the training length
(${\sim}3{,}000$ frames), beyond which prediction quality gradually
degrades.

\paragraph{Visual Odometry (VO) Mode.}
For sequences that far exceed the Direct mode's effective range
(e.g., tens of thousands of frames), we switch to VO mode.
The input is partitioned into overlapping local windows.
Within each window, an initial subset of frames is processed jointly
to establish a robust local scale and coordinate system, and the
remaining frames are processed causally via GCA with keyframe
selection.
At the end of each window, the model state is reset.
To fuse successive windows into a single global trajectory, we
compute a Sim(3) alignment between the overlapping regions of
consecutive windows, recovering the relative scale, rotation, and
translation.
This enables \method to process arbitrarily long sequences with
bounded memory, at the cost of additional drift introduced at each
window boundary: unlike Direct mode, VO mode incurs extra alignment
error that compounds with the number of windows.

\paragraph{Trade-offs.}
Direct mode produces more accurate trajectories by avoiding
inter-window alignment error, and is the preferred choice when the
sequence length stays within ${\sim}3{,}000$ frames.
For inputs that substantially exceed this range, VO mode generalizes
more effectively at the cost of accumulated alignment drift at the 
window boundaries.
In practice, the choice is determined by the sequence length and the
required level of global consistency.

\paragraph{Default Inference Configuration.}
Unless otherwise specified, all experiments in this report use Direct
Output Mode with a local pose-reference window size $k=64$ and
keyframe interval $m=1$, at a resolution of $518 \times 378$ with
bfloat16 precision.
For the large-scale demo videos that span city-scale environments
or extremely long sequences, we use VO mode.

\section{Evaluation Benchmark}
\label{sec:benchmark}

We establish a comprehensive evaluation benchmark covering 
camera pose estimation and 3D reconstruction across diverse indoor, 
outdoor, and large-scale environments.

\subsection{Datasets}
\label{sec:eval_datasets}

Streaming 3D reconstruction must generalize across a wide spectrum of
scenarios, from object-centric captures to room-scale interiors and
city-scale outdoor trajectories, under varying sequence lengths,
camera motions, and scene complexities.
To thoroughly evaluate this, our suite comprises seven
complementary benchmarks: Oxford Spires~\cite{tao2025spires},
ETH3D~\cite{schops2017multi}, 7-Scenes~\cite{shotton2013scene},
Tanks and Temples~\cite{Knapitsch2017}, NRGBD~\cite{azinovic2022neural},
Sintel~\cite{Butler:ECCV:2012}, and the KITTI odometry
benchmark~\cite{Geiger2012CVPR}. Together, they cover indoor and outdoor
environments, dynamic scenes, and trajectories from room scale to city
scale, with sequences ranging from tens to thousands of frames.
Below we describe each benchmark and our evaluation configuration.

\noindent\textbf{Oxford Spires}~\cite{tao2025spires} is a large-scale
outdoor and indoor dataset captured across the historic Oxford campus,
featuring complex scene transitions, revisits, and significant scale
variation within each sequence.
The dataset provides ground-truth camera trajectories from a
high-precision LiDAR-inertial SLAM system.
We select 10 scenes with available ground-truth trajectories: bodleian-library-02, christ-church~02--03 and~05, keble-college-02--05, observatory-quarter-01 and observatory-quarter-02.
We use camera~0 as the viewpoint and evaluate under two settings:
\textit{sparse} (320 frames, sampled every 12 frames) to test
single-pass reconstruction within our training range, and
\textit{dense} (3{,}840 frames) to stress-test long-sequence
streaming capabilities.

\noindent\textbf{ETH3D}~\cite{schops2017multi} provides
high-resolution indoor and outdoor images with ground-truth depth maps
acquired from a laser scanner.
The scenes span offices, lecture rooms, relief structures, facades,
playgrounds, terraces, and botanical gardens, covering both small-scale
indoor environments and large-scale outdoor settings.
We follow the evaluation configurations established in
DA3~\cite{depthanything3}, using all available frames.
For reconstruction evaluation, we use an F1 threshold of $d = 0.25$\,m.

\noindent\textbf{7-Scenes}~\cite{shotton2013scene} is a widely-used
indoor RGB-D dataset consisting of 7 scenes (Chess, Fire, Heads,
Office, Pumpkin, Kitchen, Stairs) captured with a Kinect sensor at
$640 \times 480$ resolution.
It is a challenging real-world dataset: the images contain significant
motion blur, and the scenes feature textureless surfaces and repetitive
structures that are difficult for pose estimation.
We downsample the number of frames for each scene by a stride of 5
to reduce redundancy from the high-framerate Kinect capture while
retaining sufficient viewpoint coverage.

\noindent\textbf{Tanks and Temples}~\cite{Knapitsch2017} is a
large-scale outdoor dataset providing high-resolution images, depth
maps from LiDAR, and ground-truth 3D shapes for benchmarking
multi-view reconstruction.
We select 6 scenes covering diverse structures: Barn, Caterpillar,
Church, Ignatius, Meetingroom, and Truck, with all images included.

\noindent\textbf{NRGBD}~\cite{azinovic2022neural} is a dataset
designed for neural RGB-D surface reconstruction, containing indoor
scenes with high-quality ground-truth depth from structured-light
sensors.
The scenes include cluttered room-scale environments with fine
geometric details.
We sample images with a stride of 5 and evaluate dense reconstruction
quality using the F1 metric.

\noindent\textbf{Sintel}~\cite{Butler:ECCV:2012} provides dynamic
sequences for evaluating camera trajectory estimation under scene
motion. We include it as a pose benchmark and report ATE in meters.

\noindent\textbf{KITTI Odometry}~\cite{Geiger2012CVPR} provides
city-scale driving trajectories for odometry evaluation. We report ATE
in meters.

\subsection{Metrics}
\label{sec:metrics}

\noindent\textbf{Camera Pose Estimation.}
We evaluate camera pose accuracy using the Area Under the Curve (AUC) of the relative pose error at angular thresholds of $3\degree$ and $30\degree$.
For trajectory-level evaluation, we report the Absolute Trajectory Error (ATE) in meters, which measures global trajectory consistency after Sim(3) alignment, as well as Relative Pose Error for translation (RPE-trans) and rotation (RPE-rot), which capture local frame-to-frame accuracy.

\noindent\textbf{3D Reconstruction.} 
We evaluate reconstruction quality using the F1 score, Accuracy (Acc), and Completeness (Comp).
The reconstructed and ground-truth point clouds are first aligned using the Umeyama~
\cite{umeyama2002least} method, followed by ICP refinement (threshold 0.1).
For ETH3D, which features large-scale scenes with relatively sparse ground-truth points, we adopt an F1 threshold of 0.25 with a voxel size of 0.039m.
For 7-Scenes and NRGBD, which contain many frames and dense depth projections producing tens of millions of points, we first downsample both point clouds to a uniform voxel grid (voxel size $4.0/512$) before ICP, and set the F1 threshold to 0.05.

\section{Experiments}

In this section, we compare \method with state-of-the-art methods for camera pose estimation and 3D reconstruction, analyze efficiency, and conduct ablation studies.
Training details and data are described in~\cref{sec:training},
the evaluation benchmark is defined in~\cref{sec:benchmark},
and inference configuration is specified in~\cref{sec:inference_mode}.

\subsection{Baseline Methods}

We compare \method against three categories of methods:

\paragraph{Offline Feed-Forward Models.}
These methods process all input frames simultaneously using
bidirectional attention, producing globally consistent reconstructions
in a single forward pass but requiring access to the full sequence.
We include VGGT~\cite{wang2025vggt},
DA3~\cite{depthanything3},
Fast3R~\cite{yang2025fast3r},
FastVGGT~\cite{shen2025fastvggt},
and $\pi^3$~\cite{wang2025pi}.

\paragraph{Optimization-Based Methods.}
These methods iteratively refine camera poses and geometry through
explicit optimization objectives such as reprojection error
minimization or bundle adjustment.
They typically achieve high accuracy but at significant computational
cost.
We include DroidSLAM~\cite{teed2021droid},
MegaSAM~\cite{li2025megasam},
and VIPE~\cite{huang2025vipe}.
We follow VGGT-SLAM~\cite{vggtslam}
to generate intrinsics for DroidSLAM~\cite{teed2021droid}.

\paragraph{Streaming Methods.}
These methods process frames causally in a streaming fashion,
predicting poses and depth without access to future frames.
This is the same setting as \method.
We include StreamVGGT~\cite{streamVGGT},
SLAM3R~\cite{liu2025slam3r},
InfiniteVGGT~\cite{yuan2026infinitevggt},
Spann3R~\cite{wang20253d},
Stream3R~\cite{stream3r2025},
CUT3R~\cite{cut3r},
TTT3R~\cite{chen2025ttt3r},
and Wint3R~\cite{li2025wint3r}.
For fair comparison, all streaming methods are evaluated without
resetting their internal state throughout each sequence.

\subsection{Camera Pose Estimation}

\paragraph{Large-Scale Trajectory Estimation on Oxford Spires.}
Oxford Spires is one of the most challenging benchmarks for streaming
pose estimation: sequences span complex indoor-outdoor environments
with abrupt scene transitions (e.g., outdoor courtyards to dark
staircases), revisits to previously observed areas after long temporal
gaps, and large scale variation across the trajectory.
These properties demand both accurate local pose estimation and
long-range global consistency, which are precisely the capabilities
GCA is designed to provide.
We evaluate under two settings to test both aspects.

\begin{figure}[p]
    \centering
    \vspace{-2em}
    \textbf{(a)} Comparison with SOTA offline and optimization-based methods on Oxford-Spires \\[2pt]
    \includegraphics[width=0.9\textwidth]{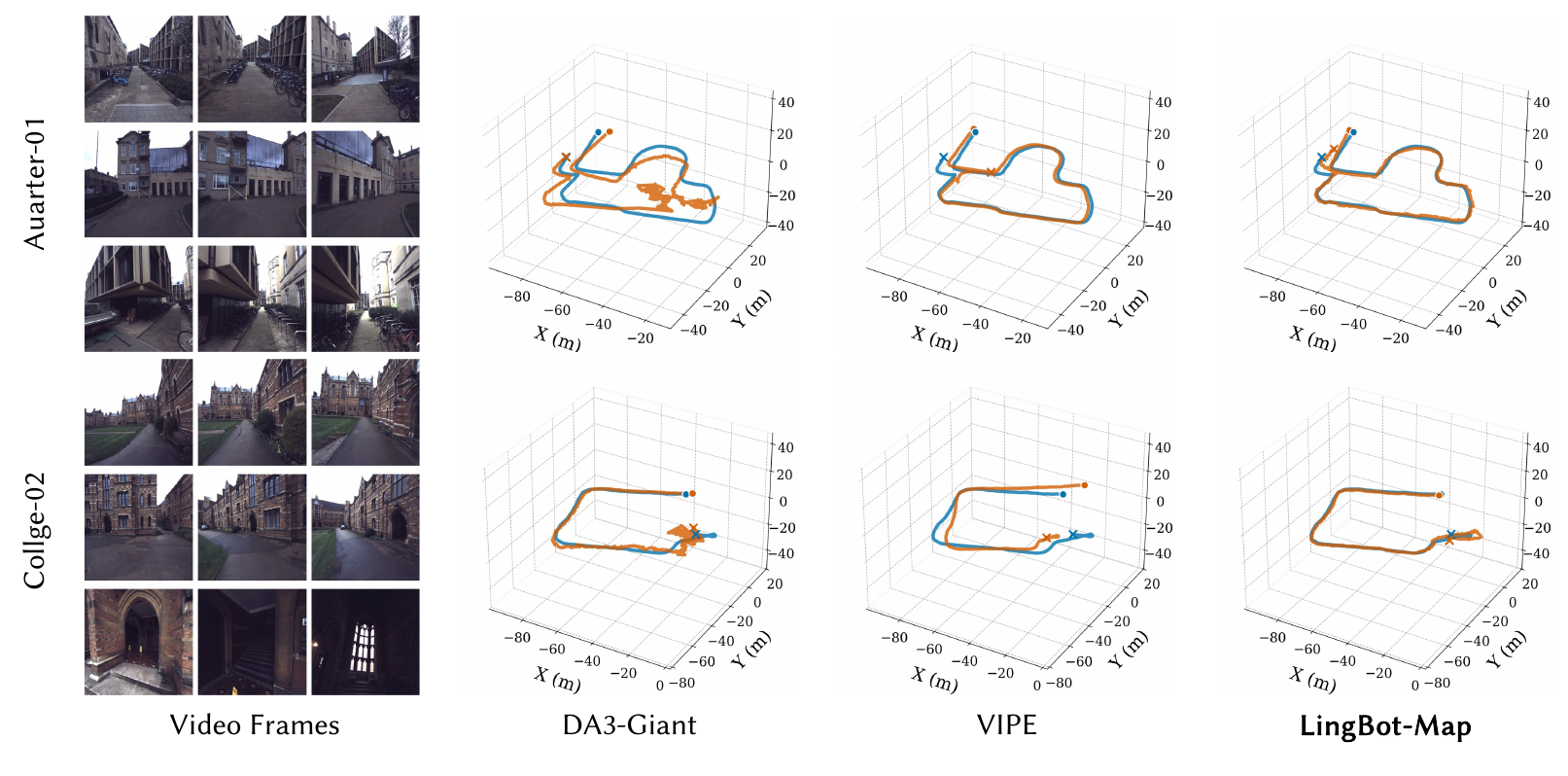}

    \vspace{3pt}

    \textbf{(b)} Comparison with SOTA streaming methods across diverse scenes \\[2pt]
    \includegraphics[width=0.9\textwidth]{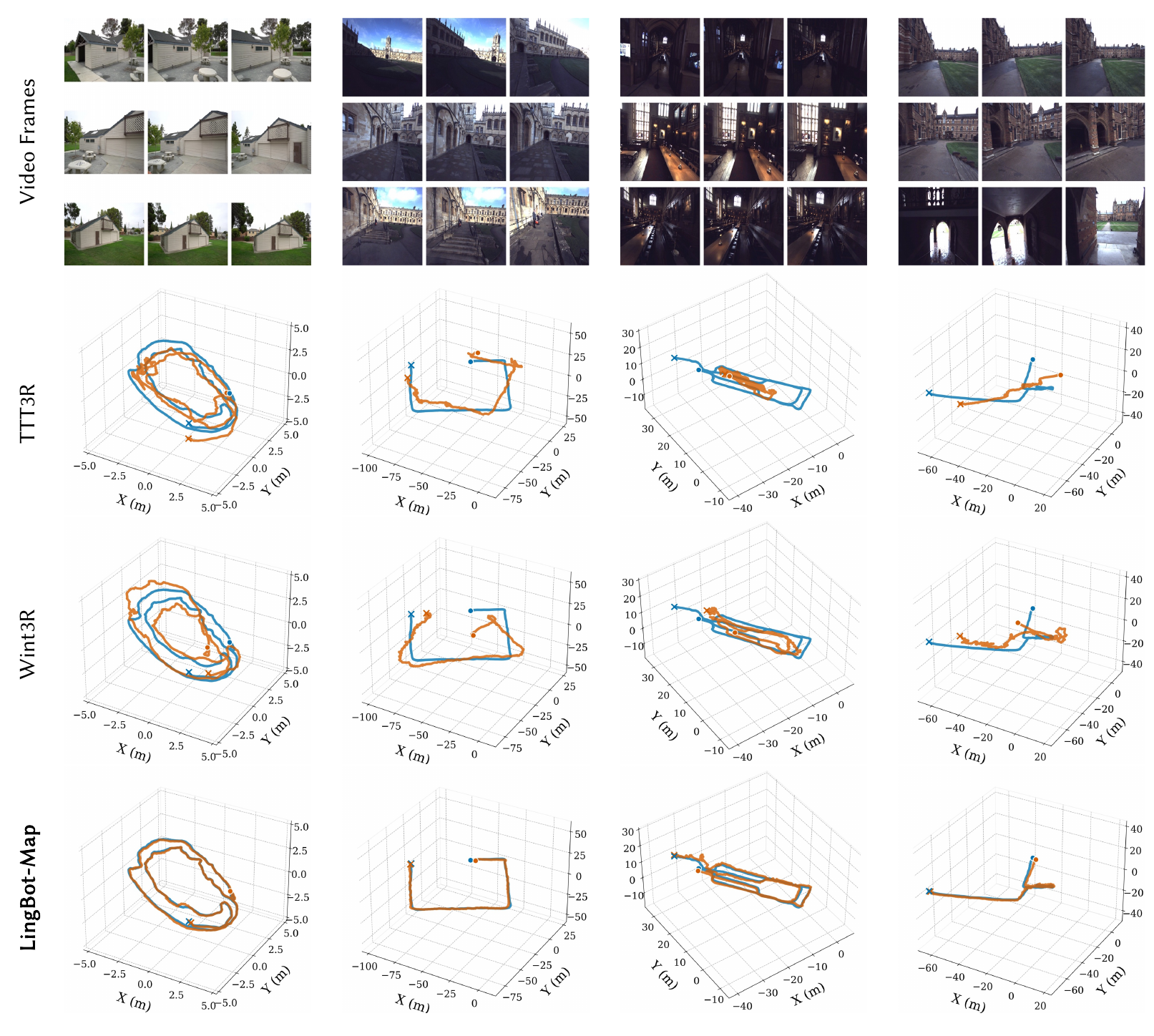}

    \definecolor{gtcolor}{RGB}{31, 119, 180}
    \definecolor{predcolor}{RGB}{255, 127, 14}

    \caption{\textbf{Trajectory comparison.}
    (a)~On Oxford-Spires, LingBot-Map evan outperforms bidirectional (DA3-Giant), and optimization-based (ViPE) methods, accurately preserving trajectories during complex outdoor-to-indoor transitions and dark staircases.
    (b)~Across Tanks and Temples and additional Oxford-Spires scenes, our method consistently produces accurate trajectories while competing streaming methods suffer from severe drift.
    \textcolor{gtcolor}{Ground truth} in blue, \textcolor{predcolor}{predicted} in orange; start~{\textbullet}, end~\textbf{\texttimes}.}
    \label{fig:traj_results}
    \vspace{-8pt}
\end{figure}

In the \textit{sparse} setting (320 frames, sampled every 12 frames),
all method categories can run on our hardware, enabling a fair
comparison across offline, optimization-based, and online approaches.
As shown in~\cref{tab:pose_accuracy_short_oxford}, \method
achieves the best results on nearly all metrics.
Despite operating in a streaming online manner, our method surpasses
the strongest offline baselines by a large margin.
\method achieves an AUC@15 of 63.22, substantially exceeding the
best offline method DA3~\cite{depthanything3} (49.84) and more than
doubling VGGT~\cite{wang2025vggt} (23.84).
On trajectory-level accuracy, \method reduces ATE from 12.87
(DA3) and 24.78 (VGGT) to 5.37.
We attribute this to a fundamental limitation of existing offline
methods: they are trained on datasets with a small number of
viewpoints where consecutive frames remain close to each other and
largely observe the same local region.
When confronted with the complex scene transitions and large
viewpoint changes in Oxford Spires, the learned priors fail to
transfer.
Compared with optimization-based methods, which explicitly minimize
reprojection error across frames, \method still achieves superior
performance.
\method outperforms the best optimization-based approach
VIPE~\cite{huang2025vipe} on both pose accuracy
(AUC@15: 63.22 vs.\ 45.35) and trajectory consistency
(ATE: 5.37 vs.\ 10.52).
VIPE relies on iterative bundle adjustment and is computationally
expensive; in contrast, \method achieves this accuracy in a single
forward pass.
Among online streaming methods, the performance gap is even more
pronounced.
All competing approaches suffer from severe memory forgetting: as
the sequence progresses, they lose track of previously observed
geometry, leading to accumulated drift.
The best online competitor, CUT3R~\cite{cut3r}, achieves an AUC@15
of only 5.98 and ATE of 18.16, while \method achieves 63.22 and
5.37 respectively, a $10.6\times$ improvement in pose accuracy and
$3.4\times$ lower trajectory error
(see~\cref{fig:traj_results}(a) for qualitative comparison).

\begin{table}[t]
\centering
\caption{\textbf{Pose and trajectory accuracy comparison on Oxford Spires (sparse setting).}
Our method achieves the best performance on most metrics among prior offline, optimization-based, and online methods. The \textbf{best} and \underline{second best} results are marked.}
\label{tab:pose_accuracy_short_oxford}
\begin{tabular}{l c ccccc}
\toprule
Methods & Type & AUC@15 $\uparrow$ & AUC@30 $\uparrow$ & ATE $\downarrow$ & RPE-trans $\downarrow$ & RPE-rot $\downarrow$ \\
\midrule
Fast3R~\cite{yang2025fast3r}      & offline & 1.20 & 2.99 & 34.80 & 8.21 & 59.51 \\
VGGT~\cite{wang2025vggt}& offline & 23.84 & 35.09& 24.78 & 8.87 & 22.79\\
DA3~\cite{depthanything3}      & offline & \underline{49.84} & \underline{56.68} & 12.87 & 3.22 & 16.17 \\
FastVGGT~\cite{shen2025fastvggt} & offline & 21.68 & 34.64 & 22.43 & 7.25 & 16.12 \\
$\pi^3$~\cite{wang2025pi} & offline & 38.64 & 48.65 & 14.03 & 2.58 & 10.33 \\
\midrule
DroidSLAM~\cite{teed2021droid}& optim & 8.58 & 21.41 & 21.84 & 1.02 & 6.90 \\
MegaSAM~\cite{li2025megasam}& optim & 15.91 & 28.03 & 13.80 & \underline{0.76} & 7.24 \\
VIPE~\cite{huang2025vipe}& optim & 45.35 & 51.88 & \underline{10.52} & \textbf{0.43} & \underline{5.98} \\
\midrule
StreamVGGT~\cite{streamVGGT}& online & 10.91 & 17.04 & 28.41 & 6.35 & 16.28 \\
SLAM3R~\cite{liu2025slam3r}& online & 1.67 & 5.10 & 29.69 & 7.57 & 27.50 \\
InfiniteVGGT~\cite{yuan2026infinitevggt} & online & 10.25 & 16.33 & 30.49 & 5.72 & 15.01 \\
Spann3R~\cite{wang20253d} & online & 2.06 & 5.09 & 32.12 & 3.54 & 14.37 \\
Stream3R-w~\cite{stream3r2025}      & online & 6.56 & 11.03 & 33.03 & 4.73 & 16.79 \\
Stream3R~\cite{stream3r2025}      & online & 9.67 & 15.21 & 29.58 & 6.67 & 16.90 \\
CUT3R~\cite{cut3r}    & online & 5.98 & 14.95 & 18.16 & 1.17 & 7.18 \\
TTT3R~\cite{chen2025ttt3r}    & online & 13.92 & 25.90 & 19.35 & 2.28 & 13.30 \\
Wint3R~\cite{li2025wint3r}      & online & 11.61 & 23.42 & 21.10 & 1.62 & 6.27 \\
\midrule
\method (Ours)  & online & \textbf{63.22} & \textbf{76.46} & \textbf{5.37} & 0.93 & \textbf{3.69} \\
\bottomrule
\end{tabular}
\end{table}

In the \textit{dense} setting (full 3,840 frames), we stress-test
long-sequence streaming capabilities
(\cref{tab:pose_accuracy_long_oxford}).
This setting is particularly revealing: as the trajectory length
increases from 320 to 3,840 frames, most feed-forward methods degrade
dramatically due to accumulated drift.
CUT3R's ATE rises from 18.16 to 32.47 ($1.8\times$ increase), and
Wint3R degrades from 21.10 to 32.90.
In contrast, \method maintains consistently low error
(5.37 $\to$ 5.60), with a marginal increase of only 0.23 over a
$12\times$ longer sequence.
This demonstrates that the three-level context structure of
GCA (anchor, trajectory memory, and local window) effectively
preserves long-range geometric consistency without explicit
optimization or loop closure.
Notably, \method achieves competitive inference speed at 20.29
FPS while maintaining the best trajectory accuracy among all
streaming methods.

\begin{table}[t]
\centering
\caption{\textbf{Sparse vs.\ dense trajectory accuracy on Oxford Spires.}
We compare ATE under the sparse (320 frames) and dense (3,840 frames) settings.
$\Delta$ATE measures the degradation from sparse to dense.
\method maintains near-constant accuracy while competing methods degrade significantly.
The \textbf{best} and \underline{second best} results are marked.}
\label{tab:pose_accuracy_long_oxford}
\setlength{\tabcolsep}{10pt}
\begin{tabular}{lcccccc}
\toprule
 & CUT3R & TTT3R & Wint3R & Inf.-VGGT & Stream3R-w & \method \\
\midrule
ATE$_{\text{sparse}}$ $\downarrow$ & \underline{18.16} & 19.35 & 21.10 & 30.49 & 33.03 & \textbf{5.37} \\
ATE$_{\text{dense}}$ $\downarrow$  & 32.47\rlap{\scriptsize{\textcolor{red}{~(+14.31)}}} & \underline{25.05}\rlap{\scriptsize{\textcolor{red}{~(+5.70)}}} & 32.90\rlap{\scriptsize{\textcolor{red}{~(+11.80)}}} & 31.75\rlap{\scriptsize{\textcolor{red}{~(+1.26)}}} & 33.73\rlap{\scriptsize{\textcolor{red}{~(+0.70)}}} & \textbf{5.60}\rlap{\scriptsize{\textcolor{green!50!black}{~(+0.23)}}} \\
\midrule
FPS $\uparrow$ & \textbf{29.21} & \underline{28.97} & 3.88 & 7.78 & 13.66 & 20.29 \\
\bottomrule
\end{tabular}
\end{table}

\paragraph{Generalization on Diverse Benchmarks.}
To verify that the strong performance on Oxford Spires is not specific
to large-scale outdoor trajectories, we evaluate on three additional
benchmarks that cover fundamentally different scales and scene types:
ETH3D~\cite{schops2017multi} (mixed indoor and outdoor scenes with
laser-scanned ground-truth depth),
7-Scenes~\cite{shotton2013scene} (room-scale RGB-D sequences with
textureless surfaces and significant motion blur),
and Tanks and Temples~\cite{Knapitsch2017} (outdoor multi-view
captures of large structures).
As shown in~\cref{tab:pose_accuracy_short}, \method consistently
outperforms all competing streaming methods by a substantial margin
across all three datasets and all metrics.

On Tanks and Temples, \method achieves an AUC@30 of 92.34 and ATE of
0.21, improving over the strongest competing results (AUC@30: 71.39
from InfiniteVGGT; ATE: 0.67 from TTT3R) by $+20.95$ AUC points and
$3.2\times$ lower ATE.
On ETH3D, our ATE of 0.43 is $2\times$ lower than the
second-best Wint3R (0.86), indicating that our model handles both the
precise indoor geometry and the broader outdoor structures in this
dataset.
On 7-Scenes, \method achieves the lowest ATE of 0.08, compared with
0.11 for the strongest competing methods, confirming robust performance
even on room-scale indoor scenes, where the main challenges are
textureless walls, repetitive structures, and heavy motion blur.
Taken together, these results demonstrate that \method is not a
specialist for any particular scenario but a general-purpose streaming
pose estimator that scales from small rooms to city-scale environments.

Qualitative trajectory comparisons are shown in~\cref{fig:traj_results}.
In part~(a), on Oxford Spires, \method accurately tracks the camera
through complex outdoor-to-indoor transitions and dark staircases,
while DA3-Giant and ViPE both exhibit significant trajectory
drift.
In part~(b), across Tanks and Temples and additional Oxford Spires
scenes, competing streaming methods (TTT3R, Wint3R) produce
trajectories that progressively diverge from the ground truth, whereas
\method consistently maintains close alignment throughout the full
sequence.

\begin{table}[t]
\centering
\caption{\textbf{Pose and trajectory accuracy comparison on ETH3D, 7-Scenes, and Tanks \& Temples.}
Results on ETH3D, 7-Scenes, and Tanks \& Temples show that our method achieves the best performance across all datasets. The \textbf{best} and \underline{second best} results are marked.}
\label{tab:pose_accuracy_short}
\resizebox{\textwidth}{!}{%
\begin{tabular}{l c ccc ccc ccc}
\toprule
\multirow{2}{*}{Methods} & \multirow{2}{*}{Type} &
\multicolumn{3}{c}{ETH3D} &
\multicolumn{3}{c}{7-Scenes} &
\multicolumn{3}{c}{Tanks \& Temples} \\
\cmidrule(lr){3-5}\cmidrule(lr){6-8}\cmidrule(lr){9-11}
 &  & Auc3 $\uparrow$ & Auc30 $\uparrow$ & ATE $\downarrow$ & Auc3 $\uparrow$ & Auc30 $\uparrow$ & ATE $\downarrow$ & Auc3 $\uparrow$ & Auc30 $\uparrow$ & ATE $\downarrow$ \\
\midrule
SLAM3R~\cite{liu2025slam3r}& online & 1.10 & 19.03 & 1.98 & 5.13 & 69.05 & \underline{0.11} & 1.68 & 37.48 & 1.42 \\
Spann3R~\cite{wang20253d} & online & 0.72 & 23.34 & 2.11 & 2.99 & 59.59 & 0.20 & 1.80 & 25.85 & 2.12 \\
InfiniteVGGT~\cite{yuan2026infinitevggt} & online & \underline{13.88} & 63.81 & 1.47 & 8.45 & 74.58 & 0.12 & 19.27 & \underline{71.39} & 1.01 \\
CUT3R~\cite{cut3r}    & online & 8.04 & 47.52 & 1.43 & 1.63 & 43.86 & 0.30 & 1.23 & 19.93 & 1.80 \\
TTT3R~\cite{chen2025ttt3r}    & online & 8.47 & 60.05 & 1.22 & 5.79 & 72.19 & \underline{0.11} & 5.42 & 64.16 & \underline{0.67} \\
Wint3R~\cite{li2025wint3r}      & online & 8.07 & \underline{64.02} & \underline{0.86} & 2.94 & 65.02 & 0.12 & 2.88 & 49.48 & 0.88 \\
Stream3R~\cite{stream3r2025}      & online & 12.41 & 60.57 & 1.67 & \underline{9.84} & \underline{74.81} & \underline{0.11} & \underline{31.88} & 71.32 & 0.76 \\
Stream3R-w~\cite{stream3r2025}      & online & 6.76 & 50.90 & 1.72 & 7.91 & 63.21 & 0.26 & 18.86 & 62.45 & 1.23 \\
\method (Ours)      & online & \textbf{40.34} & \textbf{87.96} & \textbf{0.43} & \textbf{13.20} & \textbf{79.06} & \textbf{0.08} & \textbf{44.99} & \textbf{92.34} & \textbf{0.21} \\
\bottomrule
\end{tabular}%
}
\end{table}

\begin{wraptable}[7]{r}{0.42\textwidth}
\vspace{-\baselineskip}
\centering
\captionsetup{position=top,skip=3pt}
\caption{\textbf{Comparison on Sintel and KITTI.}
We report ATE in meters; lower is better.}
\label{tab:stress}
\small
\setlength{\tabcolsep}{4pt}
\begin{tabular}{lcc}
\toprule
Method & Sintel & KITTI \\
\midrule
DROID-SLAM~\cite{teed2021droid} & 0.20 & 129.34 \\
VGGT-Long~\cite{deng2025vggt} & 0.17 & 31.89 \\
TTT3R~\cite{chen2025ttt3r} & 0.22 & 171.63 \\
\method (Ours) & \textbf{0.10} & \textbf{24.12} \\
\bottomrule
\end{tabular}
\end{wraptable}

\paragraph{Dynamic and City-Scale Stress Tests.}
We further evaluate on the dynamic Sintel~\cite{Butler:ECCV:2012}
benchmark and the city-scale KITTI~\cite{Geiger2012CVPR} odometry
benchmark. As shown in~\cref{tab:stress}, \method attains the lowest ATE
on both Sintel (0.10) and KITTI (24.12) among the evaluated methods,
despite running in a single online pass without state reset, pose-graph
optimization, or loop closure.

 \begin{figure}[t!]
    \centering
    \includegraphics[width=0.98\textwidth]{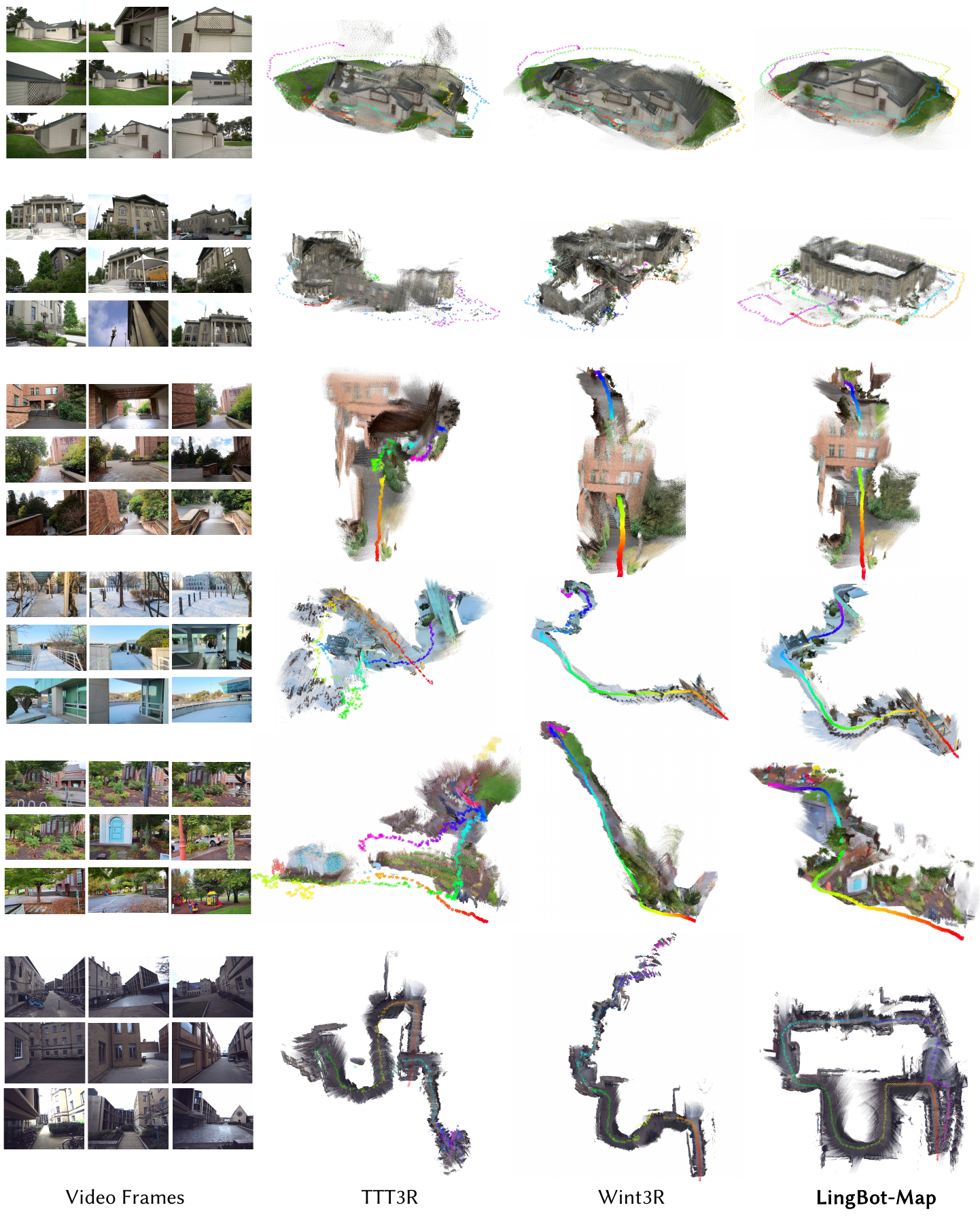}

    \caption{\textbf{Qualitative comparison of point cloud reconstruction.}
    LingBot-Map produces geometrically coherent and temporally stable reconstructions, preserving sharp building edges and continuous surfaces even over long sequences with revisits.
    TTT3R and Wint3R suffer from trajectory drift that causes fragmented geometry and duplicated structures, with degradation most severe in complex multi-building scenes (bottom rows).}
    \label{fig:point_comparison}
    \vspace{-1em}
  \end{figure}

\subsection{3D Reconstruction}

We evaluate the 3D reconstruction quality of \method on ETH3D,
7-Scenes, and NRGBD~\cite{azinovic2022neural}.
The evaluation protocol and metrics (Accuracy, Completeness, F1)
are described in~\cref{sec:metrics}.
Quantitative results are summarized in~\cref{tab:pointacc}.

Since reconstruction quality depends directly on pose accuracy
and depth estimation, the improvements reported in the previous
section translate into substantial gains in 3D reconstruction.
\method achieves the best F1 score on all three datasets.
On ETH3D, \method reaches an F1 of 86.80, outperforming the
runner-up Wint3R (77.28) by 9.52 points.
The improvement is also reflected in accuracy (0.17 vs.\ 0.28)
and completeness (0.08 vs.\ 0.21), indicating that our
reconstructions are both more precise and more complete in their
coverage of the scene.
On 7-Scenes, \method achieves the best F1 of 82.38, exceeding
Wint3R (78.81) by 3.57 points, together with the best completeness
(0.04).
The gains here are more modest because the room-scale scenes
have limited trajectory length, and most methods already perform
reasonably well; nonetheless, \method still ranks first.
On NRGBD, \method achieves an F1 of 65.10, improving over
 Wint3R (56.96) by 8.14 points.
NRGBD contains cluttered indoor environments with fine geometric
details, where accumulated pose drift in competing methods leads
to blurred or duplicated surfaces in the reconstruction.
Our drift-resistant trajectory estimation directly benefits
reconstruction fidelity in these scenarios.

A qualitative comparison is shown in~\cref{fig:point_comparison}.
In simpler scenes (top rows), all methods produce reasonable
reconstructions, but TTT3R and Wint3R already show noticeable
misalignment at building edges.
As scene complexity increases (middle rows), competing methods begin
to produce duplicated structures and blurred surfaces, a direct
consequence of accumulated pose drift causing the same geometry to
be projected to different locations.
In the most challenging multi-building outdoor scenes (bottom rows),
the gap becomes striking: TTT3R and Wint3R lose spatial coherence
entirely, producing collapsed and fragmented point clouds where
individual buildings are no longer distinguishable.
In contrast, \method maintains clean geometry with sharp structural
edges and continuous wall surfaces throughout, demonstrating that
the long-range consistency provided by GCA directly translates to
high-fidelity 3D reconstruction.

\begin{table}[t]
\centering
\caption{\textbf{Point cloud reconstruction comparison on ETH3D, 7-Scenes, and NRGBD.} Our method achieves the best F1 score on all three datasets and the best or second-best Accuracy and Completeness. The \textbf{best} and \underline{second best} results are marked.}
\label{tab:pointacc}
\resizebox{\textwidth}{!}{%
\begin{tabular}{l c ccc ccc ccc}
\toprule
\multirow{2}{*}{Methods} & \multirow{2}{*}{Type} &
\multicolumn{3}{c}{ETH3D} &
\multicolumn{3}{c}{7-Scenes} &
\multicolumn{3}{c}{NRGBD} \\
\cmidrule(lr){3-5}\cmidrule(lr){6-8}\cmidrule(lr){9-11}
 &  & Acc $\downarrow$ & Comp $\downarrow$ & F1 $\uparrow$ & Acc $\downarrow$ & Comp $\downarrow$ & F1 $\uparrow$ & Acc $\downarrow$ & Comp $\downarrow$ & F1 $\uparrow$ \\
\midrule
StreamVGGT~\cite{streamVGGT}& online & 0.64 & 0.34 & 58.11 & 0.04 & 0.11 & 69.44 & 0.13 & 0.05 & 45.08 \\
InfiniteVGGT~\cite{yuan2026infinitevggt} & online & 0.65 & 0.35 & 57.69 & 0.04 & 0.11 & 68.53 & 0.13 & 0.05 & 42.27 \\
CUT3R~\cite{cut3r}    & online & 0.57 & 0.50 & 67.63 & 0.07 & 0.10 & 58.98 & 0.25 & 0.15 & 32.22 \\
TTT3R~\cite{chen2025ttt3r} & online & 0.41 & 0.22 & 68.48 & \underline{0.03} & 0.08 & 77.25 & 0.16 & 0.06 & 53.55 \\
Wint3R~\cite{li2025wint3r}      & online & \underline{0.28} & \underline{0.21} & \underline{77.28} & \underline{0.03} & \underline{0.07} & \underline{78.81} & \underline{0.09} & \underline{0.04} & \underline{56.96} \\
Stream3R~\cite{stream3r2025}      & online & 0.44 & 0.28 & 72.87 & \textbf{0.02} & 0.09 & 78.79 & 0.21 & 0.07 & 54.07 \\
Stream3R-w~\cite{stream3r2025}      & online & 0.58 & 0.37 & 67.09 & 0.04 & 0.15 & 71.94 & 0.20 & 0.06 & 53.74 \\
\method (Ours)      & online & \textbf{0.17} & \textbf{0.08} & \textbf{86.80} & \underline{0.03} & \textbf{0.04} & \textbf{82.38} & \textbf{0.07} & \textbf{0.03} & \textbf{65.10} \\
\bottomrule
\end{tabular}%
}
\end{table}

\subsection{Ablation Study}
We conduct ablation studies to analyze the contributions of each
component in GCA.
We use TartanAir and TartanGround as the ablation testbed for their
high-quality ground-truth annotations, complex scene geometry, and
long trajectories (up to thousands of frames), which provide a
controlled yet challenging setting that isolates the effect of each
component.
We initialize the model from the first-stage checkpoint and fine-tune
on these datasets, evaluating on the TartanGround validation set
with sequences of 320 frames sampled at a stride of 8 (spanning
${\sim}2{,}560$ frames in temporal extent).
We use a learning rate of $5 \times 10^{-5}$ and follow the same
progressive view curriculum as the main streaming training
(\cref{sec:streaming_training}).
Each ablation experiment requires approximately 3,840 GPU hours.

Results are shown in~\cref{tab:ablation_long_pose}. Starting
from a baseline with only the relative pose loss (row~1), we
progressively add each component and measure its impact.

\paragraph{Anchor Initialization.}
Adding anchor initialization (row~1 $\to$ row~2) boosts AUC@3 from
9.80 to 13.63 (+3.83) and reduces ATE from 8.59 to 7.88.
Monocular streaming reconstruction is inherently scale-ambiguous:
without an explicit geometric reference, the model must implicitly
infer both the coordinate origin and the absolute scale from the
data, which becomes increasingly unreliable as the sequence grows.
The anchor context resolves this by designating the first $n$ frames
as fixed references that establish the scale and coordinate system
before streaming begins.
The improvement in AUC@3 (+3.83) confirms that the anchor mechanism
improves not just global trajectory consistency but also local pose
accuracy, as each new frame can now be registered against a
well-defined geometric reference.

\paragraph{Context Tokens (Trajectory Memory).}
Adding context tokens on top of anchor initialization (row~2 $\to$
row~4) further improves AUC@3 from 13.63 to 15.75 and reduces ATE
from 7.88 to 7.46.
The anchor context and local window together provide a fixed global
reference and dense recent observations, but without any record of
intermediate frames, pose errors accumulate unchecked between the
anchor and the current window.
Context tokens address this by retaining a compact 6-token summary
for each evicted frame, preserving the key geometric cues of the
full observation history at minimal memory cost.
The consistent improvement across all metrics (AUC@3: +2.12,
AUC@30: +1.21, ATE: $-$0.42) validates that even a lightweight
trajectory memory meaningfully reduces long-range drift.

\paragraph{Relative Pose Loss.}
Comparing row~3 (without Rel. Loss) and row~4 (with Rel. Loss)
isolates the effect of relative pose supervision.
Without it, RPE-rot degrades sharply from 2.26 to 5.35
($2.4\times$ worse) and ATE increases from 7.46 to 8.25.
The relative pose loss supervises all frame pairs within the local
pose-reference window, directly constraining the frame-to-frame
relative motion.
This is complementary to the absolute pose loss: while the absolute
loss anchors each frame's pose in the global coordinate system,
the relative loss enforces local geometric consistency within the
sliding window, preventing the small per-frame errors that
compound into trajectory drift over long sequences.
Notably, the rotation degradation ($+$3.09 in RPE-rot) is far more
severe than the translation degradation, suggesting that rotation
estimation is particularly sensitive to the lack of local
pairwise supervision.

\paragraph{Video RoPE.}
The final component, Video RoPE (row~4 $\to$ row~5), yields the
single largest ATE improvement: 7.46 $\to$ 5.98 ($-$1.48), along
with gains across all other metrics (AUC@3: +0.64, AUC@30: +1.95,
RPE-trans: $-$0.15).
Without temporal positional encoding, the trajectory memory tokens
carry geometric information but lack any notion of when each frame
was observed.
Video RoPE injects temporal ordering directly into the attention
computation, enabling the model to reason about the sequential
structure of the trajectory: how far apart two frames are in time,
and in which direction the camera has been moving.
The disproportionately large ATE improvement ($-$1.48 vs.\ the
$-$0.42 from context tokens alone) suggests that temporal ordering
is the missing ingredient that allows the trajectory memory to
realize its full potential for correcting long-range drift.

\begin{table}[t]
\centering
\caption{\textbf{Ablation study on long-sequence pose estimation and trajectory accuracy.}
All components contribute significantly to the final performance. The \textbf{best} results are marked.}
\label{tab:ablation_long_pose}
\begin{tabular}{cccc ccccc}
\toprule
Rel. Loss & A. Init. & Co. Tok. & V. RoPE & AUC@3 $\uparrow$ & AUC@30 $\uparrow$
& ATE $\downarrow$ & RPE-trans $\downarrow$ & RPE-rot $\downarrow$ \\
\midrule
\checkmark &            &            &            & 9.80  & 65.84 & 8.59 & 1.62 & 2.57 \\
\checkmark & \checkmark &            &            & 13.63 & 68.71 & 7.88 & 1.60 & 2.90 \\
           & \checkmark & \checkmark &            & 13.91 & 68.25 & 8.25 & 1.67 & 5.35 \\
\checkmark & \checkmark & \checkmark &            & 15.75 & 69.92 & 7.46 & 1.48 & 2.26 \\
\midrule
\checkmark & \checkmark & \checkmark & \checkmark & \textbf{16.39} & \textbf{71.87}
& \textbf{5.98} & \textbf{1.33} & \textbf{1.93} \\

\bottomrule
\end{tabular}
\end{table}

\paragraph{Pose-Reference Window vs.\ Full Attention.}
We compare GCA's bounded pose-reference window (size 64)
against full causal attention that retains all historical tokens
(\cref{tab:ablation_sw}).
The pose-reference window yields a $1.7\times$ speedup (11.87 $\to$
20.29 FPS) and a $2.7\times$ memory reduction (36.06 $\to$ 13.28 GB).
More importantly, the bounded window also improves trajectory
accuracy: ATE decreases from 6.60 to 5.98 and RPE-trans from 1.50
to 1.33.
This counterintuitive result can be explained by the fact that
retaining all historical image tokens introduces noise from distant,
less relevant frames that can confuse the attention computation.
GCA's design, which evicts image tokens but preserves compact
context tokens for the full trajectory, retains the essential
geometric cues while filtering out redundant information.
The only metric where full attention leads is RPE-rot (1.71 vs.\
1.93), suggesting that dense historical tokens provide slightly
richer rotational cues at the local level, but this marginal
benefit is far outweighed by the efficiency gains and the improved
global trajectory consistency.
As sequence length grows further, the gap widens: full attention's
memory and compute scale quadratically with the number of frames, while
GCA's cost remains nearly constant (6 tokens per evicted frame),
making it the only viable option for streaming at scale.

\begin{table}[t]
\centering
\caption{\textbf{Pose-reference window (size 64) vs.\ full causal attention.} The bounded window achieves better trajectory accuracy with $1.7\times$ higher speed and $2.7\times$ lower memory. The \textbf{best} and \underline{second best} results are marked.}
\label{tab:ablation_sw}
\begin{tabular}{c|ccc | cc}
\toprule
Window Size  &ATE $\downarrow$ & RPE-trans $\downarrow$ & RPE-rot $\downarrow$ & FPS$\uparrow$ & Mem (GB)$\downarrow$\\
\midrule
64        &\textbf{5.98} & \textbf{1.33}  & \underline{1.93}  &  \textbf{20.29} & \textbf{13.28}\\
Full      & \underline{6.60}  & \underline{1.50} & \textbf{1.71} & \underline{11.87} &  \underline{36.06}\\
\bottomrule
\end{tabular}
\end{table}

\paragraph{Window Size and Anchor Count.}
We additionally study the inference-time local-window size $k$ and
anchor count $n$ on Oxford Spires at $518\times378$ resolution.
As shown in~\cref{tab:window_anchor}, trajectory accuracy saturates once
$k\geq48$ and is comparatively insensitive to $n$.
We therefore use $k=64$ and $n=8$ by default, which yields an ATE of
5.3 while running at 20 FPS on an H800 and 12 FPS on an RTX 4090.

\begin{table}[t]
\centering
\caption{\textbf{Effect of inference-time window size $k$ and anchor
count $n$ on Oxford Spires.}
Each cell reports ATE (m)$\downarrow$ / H800 FPS$\uparrow$ / RTX 4090
FPS$\uparrow$ at $518\times378$ resolution.}
\label{tab:window_anchor}
\begin{tabular}{c|ccccc}
\toprule
Anchor $n$ $\backslash$ Window $k$ & 16 & 32 & 48 & 64 & 128 \\
\midrule
2 & 12.0 / 26 / 19 & 7.9 / 25 / 16 & 6.0 / 22 / 14 & 5.5 / 20 / 12 & 5.6 / 15 / 9 \\
4 & 11.9 / 26 / 19 & 7.8 / 25 / 15 & 5.8 / 22 / 14 & 5.4 / 20 / 12 & 5.4 / 15 / 9 \\
8 & 11.2 / 25 / 18 & 8.0 / 24 / 15 & 5.5 / 22 / 14 & \textbf{5.3 / 20 / 12} & 5.3 / 15 / 9 \\
\bottomrule
\end{tabular}
\end{table}

\section{Conclusion and Discussion}
\label{sec:conclusion}

We have presented \method, a streaming foundation model for long-range 3D reconstruction from continuous visual input.
At its core is Geometric Context Attention (GCA), which decomposes the streaming state into three complementary context types — anchor, local pose-reference window, and trajectory memory — inspired by the structure of classical SLAM systems but learned end-to-end.
This design reduces per-frame context growth by roughly $80{\times}$ compared to causal attention, enabling stable inference over long sequences at around 20 FPS.
Extensive evaluations across multiple benchmarks demonstrate that \method achieves state-of-the-art performance among streaming methods, and even surpasses offline and optimization-based approaches on large-scale datasets such as Oxford Spires.
By enabling accurate, real-time dense 3D reconstruction from continuous visual streams, \method opens the door to a wide range of applications, including autonomous navigation, augmented reality, and, most notably, embodied AI systems that require persistent, on-the-fly spatial understanding to interact with the physical world.

\paragraph{Limitations.}
While \method demonstrates strong performance on diverse benchmarks, several limitations remain.
First, the model currently does not incorporate explicit loop-closure detection, which could further reduce accumulated drift when revisiting previously observed regions.
Second, like other feed-forward methods, our approach does not perform test-time optimization, which could further refine reconstruction quality in challenging scenarios.
Third, extremely large-scale scenes with sequences beyond the effective range of Direct Output Mode still require window-based VO mode (\cref{sec:inference_mode}), which resets the model state between windows and aligns overlapping regions via Sim(3), introducing additional drift that accumulates as the number of windows grows.

\paragraph{Future Directions.}
A promising direction is to incorporate bundle-adjustment-like refinement and explicit loop-closure detection into the attention mechanism, further closing the gap with classical SLAM backends while retaining end-to-end differentiability.
Additionally, supporting distortion-aware camera models, extending \method to dynamic scenes with moving objects, integrating multi-modal inputs such as LiDAR or IMU data, and exploring the model as a backbone for downstream applications such as novel view synthesis and navigation are promising directions for future work.

\section*{Acknowledgements}
We thank  Jianyuan Wang, Yudong Jin, Christian Rupprecht, and Xun Cao for their helpful discussions and support.

{
\small
\bibliographystyle{plain}
\bibliography{main.bib}

@String(PAMI  = {IEEE Trans. Pattern Anal. Mach. Intell.})

@String(CVPR  = {IEEE Conf. Comput. Vis. Pattern Recog.})

@String(ICCV  = {Int. Conf. Comput. Vis.})

@String(ECCV  = {Eur. Conf. Comput. Vis.})

@String(NeurIPS = {Adv. Neural Inform. Process. Syst.})

@String(ICLR  = {Int. Conf. Learn. Represent.})

@String(CVPRW = {IEEE Conf. Comput. Vis. Pattern Recog. Worksh.})

@String(TOG   = {ACM Trans. Graph.})

@inproceedings{stream3r2025,
title={{STream3R}: Scalable Sequential {3D} Reconstruction with Causal Transformer},
author={Lan, Yushi and Luo, Yihang and Hong, Fangzhou and Zhou, Shangchen and Chen, Honghua and Lyu, Zhaoyang and Yang, Shuai and Dai, Bo and Loy, Chen Change and Pan, Xingang},
booktitle = ICLR,
year={2026}
}

@inproceedings{li2025wint3r,
title={WinT3R: Window-Based Streaming Reconstruction with Camera Token Pool}, 
author={Zizun Li and Jianjun Zhou and Yifan Wang and Haoyu Guo and Wenzheng Chang and Yang Zhou and Haoyi Zhu and Junyi Chen and Chunhua Shen and Tong He},
year={2026},
booktitle = ICLR,
}

@inproceedings{streamVGGT,
title={Streaming 4D Visual Geometry Transformer}, 
author={Dong Zhuo and Wenzhao Zheng and Jiahe Guo and Yuqi Wu and Jie Zhou and Jiwen Lu},
year={2026},
booktitle = ICLR,
}

@inproceedings{wang20253d,
  title={3d reconstruction with spatial memory},
  author={Wang, Hengyi and Agapito, Lourdes},
  booktitle={3DV},
  pages={78--89},
  year={2025},
  organization={IEEE}
}

@inproceedings{cut3r,
  Author = {Qianqian Wang* and Yifei Zhang* and Aleksander Holynski and Alexei A. Efros and Angjoo Kanazawa},
  Title = {Continuous 3D Perception Model with Persistent State},
  Year = {2025},
  booktitle=CVPR,
}

@article{chen2025ttt3r,
  title={Ttt3r: 3d reconstruction as test-time training},
  author={Chen, Xingyu and Chen, Yue and Xiu, Yuliang and Geiger, Andreas and Chen, Anpei},
  journal=ICLR,
  year={2026}
}

@inproceedings{wang2025vggt,
title={VGGT: Visual Geometry Grounded Transformer},
author={Wang, Jianyuan and Chen, Minghao and Karaev, Nikita and Vedaldi, Andrea and Rupprecht, Christian and Novotny, David},
booktitle=CVPR,
year={2025}
}

@article{yuan2026infinitevggt,
  title={InfiniteVGGT: Visual Geometry Grounded Transformer for Endless Streams},
  author={Yuan, Shuai and Yang, Yantai and Yang, Xiaotian and Zhang, Xupeng and Zhao, Zhonghao and Zhang, Lingming and Zhang, Zhipeng},
  journal={arXiv preprint arXiv:2601.02281},
  year={2026}
}

@article{depthanything3,
  title={Depth Anything 3: recovering the visual space from any views},
  author={Haotong Lin and Sili Chen and Jun Hao Liew and Donny Y. Chen and Zhenyu Li and Guang Shi and Jiashi Feng and Bingyi Kang},
  journal={arXiv preprint arXiv:2511.10647},
  year={2025}
}

@article{keetha2025mapanything,
  title={Mapanything: Universal feed-forward metric 3d reconstruction},
  author={Keetha, Nikhil and M{\"u}ller, Norman and Sch{\"o}nberger, Johannes and Porzi, Lorenzo and Zhang, Yuchen and Fischer, Tobias and Knapitsch, Arno and Zauss, Duncan and Weber, Ethan and Antunes, Nelson and others},
  journal={arXiv preprint arXiv:2509.13414},
  year={2025}
}

@inproceedings{jang2025pow3r,
  title={Pow3r: Empowering unconstrained 3d reconstruction with camera and scene priors},
  author={Jang, Wonbong and Weinzaepfel, Philippe and Leroy, Vincent and Agapito, Lourdes and Revaud, Jerome},
  booktitle=CVPR,
  pages={1071--1081},
  year={2025}
}

@inproceedings{dust3r,
  title={Dust3r: Geometric 3d vision made easy},
  author={Wang, Shuzhe and Leroy, Vincent and Cabon, Yohann and Chidlovskii, Boris and Revaud, Jerome},
  booktitle=CVPR,
  pages={20697--20709},
  year={2024}
}

@inproceedings{yang2025fast3r,
  title={Fast3r: Towards 3d reconstruction of 1000+ images in one forward pass},
  author={Yang, Jianing and Sax, Alexander and Liang, Kevin J and Henaff, Mikael and Tang, Hao and Cao, Ang and Chai, Joyce and Meier, Franziska and Feiszli, Matt},
  booktitle=CVPR,
  pages={21924--21935},
  year={2025}
}

@inproceedings{liu2025slam3r,
  title={Slam3r: Real-time dense scene reconstruction from monocular rgb videos},
  author={Liu, Yuzheng and Dong, Siyan and Wang, Shuzhe and Yin, Yingda and Yang, Yanchao and Fan, Qingnan and Chen, Baoquan},
  booktitle=CVPR,
  pages={16651--16662},
  year={2025}
}

@inproceedings{zhang2025flare,
  title={Flare: Feed-forward geometry, appearance and camera estimation from uncalibrated sparse views},
  author={Zhang, Shangzhan and Wang, Jianyuan and Xu, Yinghao and Xue, Nan and Rupprecht, Christian and Zhou, Xiaowei and Shen, Yujun and Wetzstein, Gordon},
  booktitle=CVPR,
  pages={21936--21947},
  year={2025}
}

@inproceedings{lu2025matrix3d,
  title={Matrix3d: Large photogrammetry model all-in-one},
  author={Lu, Yuanxun and Zhang, Jingyang and Fang, Tian and Nahmias, Jean-Daniel and Tsin, Yanghai and Quan, Long and Cao, Xun and Yao, Yao and Li, Shiwei},
  booktitle=CVPR,
  pages={11250--11263},
  year={2025}
}

@article{wang2025pi,
  title={$\pi^3$: Permutation-Equivariant Visual Geometry Learning},
  author={Wang, Yifan and Zhou, Jianjun and Zhu, Haoyi and Chang, Wenzheng and Zhou, Yang and Li, Zizun and Chen, Junyi and Pang, Jiangmiao and Shen, Chunhua and He, Tong},
  journal={arXiv preprint arXiv:2507.13347},
  year={2025}
}

@article{shen2025fastvggt,
  title={Fastvggt: Training-free acceleration of visual geometry transformer},
  author={Shen, You and Zhang, Zhipeng and Qu, Yansong and Zheng, Xiawu and Ji, Jiayi and Zhang, Shengchuan and Cao, Liujuan},
  journal={arXiv preprint arXiv:2509.02560},
  year={2025}
}

@inproceedings{zhang2024monst3r,
  title={MonST3R: A Simple Approach for Estimating Geometry in the Presence of Motion},
  author={Zhang, Junyi and Herrmann, Charles and Hur, Junhwa and Jampani, Varun and Darrell, Trevor and Cole, Forrester and Sun, Deqing and Yang, Ming-Hsuan},
  booktitle=ICLR,
  year={2025}
}

@inproceedings{sucar2025dynamic,
  title={Dynamic point maps: A versatile representation for dynamic 3d reconstruction},
  author={Sucar, Edgar and Lai, Zihang and Insafutdinov, Eldar and Vedaldi, Andrea},
  booktitle=ICCV,
  pages={7295--7305},
  year={2025}
}

@inproceedings{chen2025easi3r,
  title={Easi3r: Estimating disentangled motion from dust3r without training},
  author={Chen, Xingyu and Chen, Yue and Xiu, Yuliang and Geiger, Andreas and Chen, Anpei},
  booktitle=ICCV,
  pages={9158--9168},
  year={2025}
}

@inproceedings{feng2025st4rtrack,
  title={St4rtrack: Simultaneous 4d reconstruction and tracking in the world},
  author={Feng, Haiwen and Zhang, Junyi and Wang, Qianqian and Ye, Yufei and Yu, Pengcheng and Black, Michael J and Darrell, Trevor and Kanazawa, Angjoo},
  booktitle=ICCV,
  pages={8503--8513},
  year={2025}
}

@inproceedings{xiao2025spatialtracker,
  title={SpatialTrackerV2: 3D Point Tracking Made Easy},
  author={Xiao, Yuxi and Wang, Jianyuan and Xue, Nan and Karaev, Nikita and Makarov, Iurii and Kang, Bingyi and Zhu, Xin and Bao, Hujun and Shen, Yujun and Zhou, Xiaowei},
  booktitle=ICCV,
  year={2025}
}

@inproceedings{lu2025align3r,
  title={Align3r: Aligned monocular depth estimation for dynamic videos},
  author={Lu, Jiahao and Huang, Tianyu and Li, Peng and Dou, Zhiyang and Lin, Cheng and Cui, Zhiming and Dong, Zhen and Yeung, Sai-Kit and Wang, Wenping and Liu, Yuan},
  booktitle={Proceedings of the Computer Vision and Pattern Recognition Conference},
  pages={22820--22830},
  year={2025}
}

@inproceedings{jin2025stereo4d,
  title={{Stereo4D: Learning How Things Move in 3D from Internet Stereo Videos}}, 
  author={Jin, Linyi and Tucker, Richard and Li, Zhengqi and Fouhey, David and Snavely, Noah and Holynski, Aleksander},
  booktitle=CVPR,
  year={2025},
}

@inproceedings{mast3rslam,
  title={Mast3r-slam: Real-time dense slam with 3d reconstruction priors},
  author={Murai, Riku and Dexheimer, Eric and Davison, Andrew J},
  booktitle=CVPR,
  pages={16695--16705},
  year={2025}
}

@article{vggtslam,
  title={VGGT-SLAM: Dense RGB SLAM Optimized on the SL (4) Manifold},
  author={Maggio, Dominic and Lim, Hyungtae and Carlone, Luca},
  journal=NeurIPS,
  volume={39},
  year={2025}
}

@inproceedings{li2025megasam,
  title={Megasam: Accurate, fast and robust structure and motion from casual dynamic videos},
  author={Li, Zhengqi and Tucker, Richard and Cole, Forrester and Wang, Qianqian and Jin, Linyi and Ye, Vickie and Kanazawa, Angjoo and Holynski, Aleksander and Snavely, Noah},
  booktitle=CVPR,
  pages={10486--10496},
  year={2025}
}

@article{huang2025vipe,
  title={Vipe: Video pose engine for 3d geometric perception},
  author={Huang, Jiahui and Zhou, Qunjie and Rabeti, Hesam and Korovko, Aleksandr and Ling, Huan and Ren, Xuanchi and Shen, Tianchang and Gao, Jun and Slepichev, Dmitry and Lin, Chen-Hsuan and others},
  journal={arXiv preprint arXiv:2508.10934},
  year={2025}
}

@article{deng2025vggt,
  title={VGGT-Long: Chunk it, Loop it, Align it--Pushing VGGT's Limits on Kilometer-scale Long RGB Sequences},
  author={Deng, Kai and Ti, Zexin and Xu, Jiawei and Yang, Jian and Xie, Jin},
  journal={arXiv preprint arXiv:2507.16443},
  year={2025}
}

@article{teed2021droid,
  title={Droid-slam: Deep visual slam for monocular, stereo, and rgb-d cameras},
  author={Teed, Zachary and Deng, Jia},
  journal=NeurIPS,
  volume={34},
  pages={16558--16569},
  year={2021}
}

@inproceedings{lin2025longsplat,
  title={Longsplat: Robust unposed 3d gaussian splatting for casual long videos},
  author={Lin, Chin-Yang and Sun, Cheng and Yang, Fu-En and Chen, Min-Hung and Lin, Yen-Yu and Liu, Yu-Lun},
  booktitle=ICCV,
  pages={27412--27422},
  year={2025}
}

@inproceedings{ye2024no,
  title={No pose, no problem: Surprisingly simple 3d gaussian splats from sparse unposed images},
  author={Ye, Botao and Liu, Sifei and Xu, Haofei and Li, Xueting and Pollefeys, Marc and Yang, Ming-Hsuan and Peng, Songyou},
  booktitle=ICLR,
  year={2025}
}

@article{jiang2025anysplat,
  title={Anysplat: Feed-forward 3d gaussian splatting from unconstrained views},
  author={Jiang, Lihan and Mao, Yucheng and Xu, Linning and Lu, Tao and Ren, Kerui and Jin, Yichen and Xu, Xudong and Yu, Mulin and Pang, Jiangmiao and Zhao, Feng and others},
  journal=TOG,
  volume={44},
  number={6},
  pages={1--16},
  year={2025},
  publisher={ACM New York, NY, USA}
}

@inproceedings{chen2025long3r,
  title={Long3r: Long sequence streaming 3d reconstruction},
  author={Chen, Zhuoguang and Qin, Minghui and Yuan, Tianyuan and Liu, Zhe and Zhao, Hang},
  booktitle=ICCV,
  pages={5273--5284},
  year={2025}
}

@article{smart2024splatt3r,
  title={Splatt3r: Zero-shot gaussian splatting from uncalibrated image pairs},
  author={Smart, Brandon and Zheng, Chuanxia and Laina, Iro and Prisacariu, Victor Adrian},
  journal={arXiv preprint arXiv:2408.13912},
  year={2024}
}

@article{jacobs2023deepspeed,
  title={Deepspeed ulysses: System optimizations for enabling training of extreme long sequence transformer models},
  author={Jacobs, Sam Ade and Tanaka, Masahiro and Zhang, Chengming and Zhang, Minjia and Song, Shuaiwen Leon and Rajbhandari, Samyam and He, Yuxiong},
  journal={arXiv preprint arXiv:2309.14509},
  year={2023}
}

@article{wan2025,
      title={Wan: Open and Advanced Large-Scale Video Generative Models}, 
      author={Team Wan and Ang Wang and Baole Ai and Bin Wen and Chaojie Mao and Chen-Wei Xie and Di Chen and Feiwu Yu and Haiming Zhao and Jianxiao Yang and Jianyuan Zeng and Jiayu Wang and Jingfeng Zhang and Jingren Zhou and Jinkai Wang and Jixuan Chen and Kai Zhu and Kang Zhao and Keyu Yan and Lianghua Huang and Mengyang Feng and Ningyi Zhang and Pandeng Li and Pingyu Wu and Ruihang Chu and Ruili Feng and Shiwei Zhang and Siyang Sun and Tao Fang and Tianxing Wang and Tianyi Gui and Tingyu Weng and Tong Shen and Wei Lin and Wei Wang and Wei Wang and Wenmeng Zhou and Wente Wang and Wenting Shen and Wenyuan Yu and Xianzhong Shi and Xiaoming Huang and Xin Xu and Yan Kou and Yangyu Lv and Yifei Li and Yijing Liu and Yiming Wang and Yingya Zhang and Yitong Huang and Yong Li and You Wu and Yu Liu and Yulin Pan and Yun Zheng and Yuntao Hong and Yupeng Shi and Yutong Feng and Zeyinzi Jiang and Zhen Han and Zhi-Fan Wu and Ziyu Liu},
      journal = {arXiv preprint arXiv:2503.20314},
      year={2025}
}

@misc{oquab2023dinov2,
  title={DINOv2: Learning Robust Visual Features without Supervision},
  author={Oquab, Maxime and Darcet, Timothée and Moutakanni, Theo and Vo, Huy V. and Szafraniec, Marc and Khalidov, Vasil and Fernandez, Pierre and Haziza, Daniel and Massa, Francisco and El-Nouby, Alaaeldin and Howes, Russell and Huang, Po-Yao and Xu, Hu and Sharma, Vasu and Li, Shang-Wen and Galuba, Wojciech and Rabbat, Mike and Assran, Mido and Ballas, Nicolas and Synnaeve, Gabriel and Misra, Ishan and Jegou, Herve and Mairal, Julien and Labatut, Patrick and Joulin, Armand and Bojanowski, Piotr},
  journal={arXiv:2304.07193},
  year={2023}
}

@article{ye2025flashinfer,
    title = {FlashInfer: Efficient and Customizable Attention Engine for LLM Inference Serving},
    author = {
      Ye, Zihao and
      Chen, Lequn and
      Lai, Ruihang and
      Lin, Wuwei and
      Zhang, Yineng and
      Wang, Stephanie and
      Chen, Tianqi and
      Kasikci, Baris and
      Grover, Vinod and
      Krishnamurthy, Arvind and
      Ceze, Luis
    },
    journal = {arXiv preprint arXiv:2501.01005},
    year = {2025},
    url = {https://arxiv.org/abs/2501.01005}
}

@inproceedings{kwon2023efficient,
  title={Efficient Memory Management for Large Language Model Serving with PagedAttention},
  author={Woosuk Kwon and Zhuohan Li and Siyuan Zhuang and Ying Sheng and Lianmin Zheng and Cody Hao Yu and Joseph E. Gonzalez and Hao Zhang and Ion Stoica},
  booktitle={Proceedings of the ACM SIGOPS 29th Symposium on Operating Systems Principles},
  year={2023}
}

@inproceedings{wang2025spatialvid,
  title={Spatialvid: A large-scale video dataset with spatial annotations},
  author={Wang, Jiahao and Yuan, Yufeng and Zheng, Rujie and Lin, Youtian and Gao, Jian and Chen, Lin-Zhuo and Bao, Yajie and Zhang, Yi and Zeng, Chang and Zhou, Yanxi and others},
  booktitle=CVPR,
  year={2026}
}

@inproceedings{reizenstein2021common,
  title={Common objects in 3d: Large-scale learning and evaluation of real-life 3d category reconstruction},
  author={Reizenstein, Jeremy and Shapovalov, Roman and Henzler, Philipp and Sbordone, Luca and Labatut, Patrick and Novotny, David},
  booktitle=ICCV,
  pages={10901--10911},
  year={2021}
}

@inproceedings{yao2020blendedmvs,
  title={Blendedmvs: A large-scale dataset for generalized multi-view stereo networks},
  author={Yao, Yao and Luo, Zixin and Li, Shiwei and Zhang, Jingyang and Ren, Yufan and Zhou, Lei and Fang, Tian and Quan, Long},
  booktitle=CVPR,
  pages={1790--1799},
  year={2020}
}

@inproceedings{ling2024dl3dv,
  title={Dl3dv-10k: A large-scale scene dataset for deep learning-based 3d vision},
  author={Ling, Lu and Sheng, Yichen and Tu, Zhi and Zhao, Wentian and Xin, Cheng and Wan, Kun and Yu, Lantao and Guo, Qianyu and Yu, Zixun and Lu, Yawen and others},
  booktitle=CVPR,
  pages={22160--22169},
  year={2024}
}

@inproceedings{li2018megadepth,
  title={Megadepth: Learning single-view depth prediction from internet photos},
  author={Li, Zhengqi and Snavely, Noah},
  booktitle=CVPR,
  pages={2041--2050},
  year={2018}
}

@inproceedings{greff2022kubric,
  title={Kubric: A scalable dataset generator},
  author={Greff, Klaus and Belletti, Francois and Beyer, Lucas and Doersch, Carl and Du, Yilun and Duckworth, Daniel and Fleet, David J and Gnanapragasam, Dan and Golemo, Florian and Herrmann, Charles and others},
  booktitle=CVPR,
  pages={3749--3761},
  year={2022}
}

@inproceedings{xia2024rgbd,
  title={Rgbd objects in the wild: Scaling real-world 3d object learning from rgb-d videos},
  author={Xia, Hongchi and Fu, Yang and Liu, Sifei and Wang, Xiaolong},
  booktitle=CVPR,
  pages={22378--22389},
  year={2024}
}

@inproceedings{dai2017scannet,
  title={Scannet: Richly-annotated 3d reconstructions of indoor scenes},
  author={Dai, Angela and Chang, Angel X and Savva, Manolis and Halber, Maciej and Funkhouser, Thomas and Nie{\ss}ner, Matthias},
  booktitle=CVPR,
  pages={5828--5839},
  year={2017}
}

@inproceedings{roberts2021hypersim,
  title={Hypersim: A photorealistic synthetic dataset for holistic indoor scene understanding},
  author={Roberts, Mike and Ramapuram, Jason and Ranjan, Anurag and Kumar, Atulit and Bautista, Miguel Angel and Paczan, Nathan and Webb, Russ and Susskind, Joshua M},
  booktitle=ICCV,
  pages={10912--10922},
  year={2021}
}

@inproceedings{huang2018deepmvs,
  title={Deepmvs: Learning multi-view stereopsis},
  author={Huang, Po-Han and Matzen, Kevin and Kopf, Johannes and Ahuja, Narendra and Huang, Jia-Bin},
  booktitle=CVPR,
  pages={2821--2830},
  year={2018}
}

@inproceedings{zheng2023pointodyssey,
  title={Pointodyssey: A large-scale synthetic dataset for long-term point tracking},
  author={Zheng, Yang and Harley, Adam W and Shen, Bokui and Wetzstein, Gordon and Guibas, Leonidas J},
  booktitle=ICCV,
  pages={19855--19865},
  year={2023}
}

@article{cabon2020virtual,
  title={Virtual kitti 2},
  author={Cabon, Yohann and Murray, Naila and Humenberger, Martin},
  journal={arXiv preprint arXiv:2001.10773},
  year={2020}
}

@inproceedings{pan2023aria,
  title={Aria digital twin: A new benchmark dataset for egocentric 3d machine perception},
  author={Pan, Xiaqing and Charron, Nicholas and Yang, Yongqian and Peters, Scott and Whelan, Thomas and Kong, Chen and Parkhi, Omkar and Newcombe, Richard and Ren, Yuheng Carl},
  booktitle=ICCV,
  pages={20133--20143},
  year={2023}
}

@inproceedings{deitke2023objaverse,
  title={Objaverse: A universe of annotated 3d objects},
  author={Deitke, Matt and Schwenk, Dustin and Salvador, Jordi and Weihs, Luca and Michel, Oscar and VanderBilt, Eli and Schmidt, Ludwig and Ehsani, Kiana and Kembhavi, Aniruddha and Farhadi, Ali},
  booktitle=CVPR,
  pages={13142--13153},
  year={2023}
}

@inproceedings{li2023matrixcity,
  title={Matrixcity: A large-scale city dataset for city-scale neural rendering and beyond},
  author={Li, Yixuan and Jiang, Lihan and Xu, Linning and Xiangli, Yuanbo and Wang, Zhenzhi and Lin, Dahua and Dai, Bo},
  booktitle=ICCV,
  pages={3205--3215},
  year={2023}
}

@article{liao2022kitti,
  title={Kitti-360: A novel dataset and benchmarks for urban scene understanding in 2d and 3d},
  author={Liao, Yiyi and Xie, Jun and Geiger, Andreas},
  journal=PAMI,
  volume={45},
  number={3},
  pages={3292--3310},
  year={2022},
  publisher={IEEE}
}

@inproceedings{azinovic2022neural,
  title={Neural rgb-d surface reconstruction},
  author={Azinovi{\'c}, Dejan and Martin-Brualla, Ricardo and Goldman, Dan B and Nie{\ss}ner, Matthias and Thies, Justus},
  booktitle={Proceedings of the IEEE/CVF conference on computer vision and pattern recognition},
  pages={6290--6301},
  year={2022}
}

@inproceedings{wang2020tartanair,
  title={Tartanair: A dataset to push the limits of visual slam},
  author={Wang, Wenshan and Zhu, Delong and Wang, Xiangwei and Hu, Yaoyu and Qiu, Yuheng and Wang, Chen and Hu, Yafei and Kapoor, Ashish and Scherer, Sebastian},
  booktitle={2020 IEEE/RSJ International Conference on Intelligent Robots and Systems (IROS)},
  pages={4909--4916},
  year={2020},
  organization={IEEE}
}

@inproceedings{patel2025tartanground,
  title={Tartanground: A large-scale dataset for ground robot perception and navigation},
  author={Patel, Manthan and Yang, Fan and Qiu, Yuheng and Cadena, Cesar and Scherer, Sebastian and Hutter, Marco and Wang, Wenshan},
  booktitle={2025 IEEE/RSJ International Conference on Intelligent Robots and Systems (IROS)},
  pages={20524--20531},
  year={2025},
  organization={IEEE}
}

@INPROCEEDINGS{Fonder2019MidAir,
author = {Michael Fonder and Marc Van Droogenbroeck},
title = {Mid-Air: A multi-modal dataset for extremely low altitude drone flights},
booktitle = {Conference on Computer Vision and Pattern Recognition Workshop (CVPRW)},
year = {2019},
month = {June}
}

@inproceedings{tosi2021smd,
  title={Smd-nets: Stereo mixture density networks},
  author={Tosi, Fabio and Liao, Yiyi and Schmitt, Carolin and Geiger, Andreas},
  booktitle=CVPR,
  pages={8942--8952},
  year={2021}
}

@inproceedings{arnold2022map,
  title={Map-free visual relocalization: Metric pose relative to a single image},
  author={Arnold, Eduardo and Wynn, Jamie and Vicente, Sara and Garcia-Hernando, Guillermo and Monszpart, Aron and Prisacariu, Victor and Turmukhambetov, Daniyar and Brachmann, Eric},
  booktitle=ECCV,
  pages={690--708},
  year={2022},
  organization={Springer}
}

@inproceedings{sun2020scalability,
  title={Scalability in perception for autonomous driving: Waymo open dataset},
  author={Sun, Pei and Kretzschmar, Henrik and Dotiwalla, Xerxes and Chouard, Aurelien and Patnaik, Vijaysai and Tsui, Paul and Guo, James and Zhou, Yin and Chai, Yuning and Caine, Benjamin and others},
  booktitle=CVPR,
  pages={2446--2454},
  year={2020}
}

@article{zhang2025texverse,
  title={Texverse: A universe of 3d objects with high-resolution textures},
  author={Zhang, Yibo and Zhang, Li and Ma, Rui and Cao, Nan},
  journal={arXiv preprint arXiv:2508.10868},
  year={2025}
}

@inproceedings{yeshwanth2023scannet++,
  title={Scannet++: A high-fidelity dataset of 3d indoor scenes},
  author={Yeshwanth, Chandan and Liu, Yueh-Cheng and Nie{\ss}ner, Matthias and Dai, Angela},
  booktitle=ICCV,
  pages={12--22},
  year={2023}
}

@inproceedings{shotton2013scene,
  title={Scene coordinate regression forests for camera relocalization in RGB-D images},
  author={Shotton, Jamie and Glocker, Ben and Zach, Christopher and Izadi, Shahram and Criminisi, Antonio and Fitzgibbon, Andrew},
  booktitle=CVPR,
  pages={2930--2937},
  year={2013}
}

@inproceedings{schops2017multi,
  title={A multi-view stereo benchmark with high-resolution images and multi-camera videos},
  author={Schops, Thomas and Schonberger, Johannes L and Galliani, Silvano and Sattler, Torsten and Schindler, Konrad and Pollefeys, Marc and Geiger, Andreas},
  booktitle=CVPR,
  pages={3260--3269},
  year={2017}
}

@article{tao2025spires,
title={The Oxford Spires Dataset: Benchmarking Large-Scale LiDAR-Visual Localisation, Reconstruction and Radiance Field Methods},
author={Tao, Yifu and Mu{\~n}oz-Ba{\~n}{\'o}n, Miguel {\'A}ngel and Zhang, Lintong and Wang, Jiahao and Fu, Lanke Frank Tarimo and Fallon, Maurice},
journal={International Journal of Robotics Research}, 
year={2025},
}

@article{Knapitsch2017,
    author    = {Arno Knapitsch and Jaesik Park and Qian-Yi Zhou and Vladlen Koltun},
    title     = {Tanks and Temples: Benchmarking Large-Scale Scene Reconstruction},
    journal   = TOG,
    volume    = {36},
    number    = {4},
    year      = {2017},
}

@article{wang2020flow,
  title={Flow-motion and depth network for monocular stereo and beyond},
  author={Wang, Kaixuan and Shen, Shaojie},
  journal={IEEE Robotics and Automation Letters},
  volume={5},
  number={2},
  pages={3307--3314},
  year={2020},
  publisher={IEEE}
}

@inproceedings{McCormac:etal:ICCV2017,
author = {John McCormac and
      Ankur Handa and
          Stefan Leutenegger and
          Andrew J.Davison},
title = {SceneNet RGB-D: Can 5M Synthetic Images Beat Generic ImageNet Pre-training on Indoor Segmentation?},
booktitle = ICCV,
year = {2017}
}

@article{snavely2006photo,
  author       = {Noah Snavely and
                  Steven M. Seitz and
                  Richard Szeliski},
  title        = {Photo tourism: exploring photo collections in 3D},
  journal      = TOG,
  volume       = {25},
  number       = {3},
  pages        = {835--846},
  year         = {2006},
}

@inproceedings{schonberger2016structure,
  title={Structure-from-motion revisited},
  author={Schonberger, Johannes L and Frahm, Jan-Michael},
  booktitle={Proceedings of the IEEE conference on computer vision and pattern recognition},
  pages={4104--4113},
  year={2016}
}

@inproceedings{pan2024global,
  title={Global structure-from-motion revisited},
  author={Pan, Linfei and Bar{\'a}th, D{\'a}niel and Pollefeys, Marc and Sch{\"o}nberger, Johannes L},
  booktitle={European Conference on Computer Vision},
  pages={58--77},
  year={2024},
  organization={Springer}
}

@article{mur2015orb,
  title={ORB-SLAM: A versatile and accurate monocular SLAM system},
  author={Mur-Artal, Raul and Montiel, Jose Maria Martinez and Tardos, Juan D},
  journal={IEEE transactions on robotics},
  volume={31},
  number={5},
  pages={1147--1163},
  year={2015},
  publisher={IEEE}
}

@article{mur2017orb,
  title={Orb-slam2: An open-source slam system for monocular, stereo, and rgb-d cameras},
  author={Mur-Artal, Raul and Tard{\'o}s, Juan D},
  journal={IEEE transactions on robotics},
  volume={33},
  number={5},
  pages={1255--1262},
  year={2017},
  publisher={IEEE}
}

@article{campos2021orb,
  title={Orb-slam3: An accurate open-source library for visual, visual--inertial, and multimap slam},
  author={Campos, Carlos and Elvira, Richard and Rodr{\'\i}guez, Juan J G{\'o}mez and Montiel, Jos{\'e} MM and Tard{\'o}s, Juan D},
  journal={IEEE transactions on robotics},
  volume={37},
  number={6},
  pages={1874--1890},
  year={2021},
  publisher={IEEE}
}

@article{furukawa2009accurate,
  title={Accurate, dense, and robust multiview stereopsis},
  author={Furukawa, Yasutaka and Ponce, Jean},
  journal={IEEE transactions on pattern analysis and machine intelligence},
  volume={32},
  number={8},
  pages={1362--1376},
  year={2009},
  publisher={IEEE}
}

@inproceedings{schonberger2016pixelwise,
  title={Pixelwise view selection for unstructured multi-view stereo},
  author={Sch{\"o}nberger, Johannes L and Zheng, Enliang and Frahm, Jan-Michael and Pollefeys, Marc},
  booktitle={European conference on computer vision},
  pages={501--518},
  year={2016},
  organization={Springer}
}

@inproceedings{detone2018superpoint,
  title={Superpoint: Self-supervised interest point detection and description},
  author={DeTone, Daniel and Malisiewicz, Tomasz and Rabinovich, Andrew},
  booktitle={Proceedings of the IEEE conference on computer vision and pattern recognition workshops},
  pages={224--236},
  year={2018}
}

@inproceedings{sarlin2020superglue,
  title={Superglue: Learning feature matching with graph neural networks},
  author={Sarlin, Paul-Edouard and DeTone, Daniel and Malisiewicz, Tomasz and Rabinovich, Andrew},
  booktitle={Proceedings of the IEEE/CVF conference on computer vision and pattern recognition},
  pages={4938--4947},
  year={2020}
}

@inproceedings{sun2021loftr,
  title={LoFTR: Detector-free local feature matching with transformers},
  author={Sun, Jiaming and Shen, Zehong and Wang, Yuang and Bao, Hujun and Zhou, Xiaowei},
  booktitle={Proceedings of the IEEE/CVF conference on computer vision and pattern recognition},
  pages={8922--8931},
  year={2021}
}

@inproceedings{yao2018mvsnet,
  title={Mvsnet: Depth inference for unstructured multi-view stereo},
  author={Yao, Yao and Luo, Zixin and Li, Shiwei and Fang, Tian and Quan, Long},
  booktitle={Proceedings of the European conference on computer vision (ECCV)},
  pages={767--783},
  year={2018}
}

@inproceedings{yao2019recurrent,
  title={Recurrent mvsnet for high-resolution multi-view stereo depth inference},
  author={Yao, Yao and Luo, Zixin and Li, Shiwei and Shen, Tianwei and Fang, Tian and Quan, Long},
  booktitle={Proceedings of the IEEE/CVF conference on computer vision and pattern recognition},
  pages={5525--5534},
  year={2019}
}

@inproceedings{wang2024vggsfm,
  title={Vggsfm: Visual geometry grounded deep structure from motion},
  author={Wang, Jianyuan and Karaev, Nikita and Rupprecht, Christian and Novotny, David},
  booktitle={Proceedings of the IEEE/CVF conference on computer vision and pattern recognition},
  pages={21686--21697},
  year={2024}
}

@inproceedings{
   liang2025torchtitan,
   title={TorchTitan: One-stop PyTorch native solution for production ready {LLM} pretraining},
   author={Wanchao Liang and Tianyu Liu and Less Wright and Will Constable and Andrew Gu and Chien-Chin Huang and Iris Zhang and Wei Feng and Howard Huang and Junjie Wang and Sanket Purandare and Gokul Nadathur and Stratos Idreos},
   booktitle=ICLR,
   year={2025},
   url={https://openreview.net/forum?id=SFN6Wm7YBI}
}

@inproceedings{Geiger2012CVPR,
  author = {Andreas Geiger and Philip Lenz and Raquel Urtasun},
  title = {Are we ready for Autonomous Driving? The KITTI Vision Benchmark Suite},
  booktitle = CVPR,
  year = {2012}
}

@inproceedings{Butler:ECCV:2012,
title = {A naturalistic open source movie for optical flow evaluation},
author = {Butler, D. J. and Wulff, J. and Stanley, G. B. and Black, M. J.},
booktitle = ECCV,
editor = {{A. Fitzgibbon et al. (Eds.)}},
publisher = {Springer-Verlag},
series = {Part IV, LNCS 7577},
month = oct,
pages = {611--625},
year = {2012}
}

@article{zhang2026loger,
   title={LoGeR: Long-Context Geometric Reconstruction with Hybrid Memory},
   author={Junyi Zhang and Charles Herrmann and Junhwa Hur and Chen Sun and Ming-Hsuan Yang and Forrester Cole and Trevor Darrell and Deqing Sun},
   journal={arXiv preprint arXiv:2603.03269},
   year={2026}
}

@inproceedings{xie2026scal3r,
   title={Scal3R: Scalable Test-Time Training for Large-Scale 3D Reconstruction},
   author={Tao Xie and Peishan Yang and Yudong Jin and Yingfeng Cai and Wei Yin and Weiqiang Ren and Qian Zhang and Wei Hua and Sida Peng and Xiaoyang Guo and Xiaowei Zhou},
   booktitle=CVPR,
   year={2026}
}

@inproceedings{xu2024grm,
   title={{GRM}: Large Gaussian Reconstruction Model for Efficient 3D Reconstruction and Generation},
   author={Yinghao Xu and Zifan Shi and Wang Yifan and Hansheng Chen and Ceyuan Yang and Sida Peng and Yujun Shen and Gordon Wetzstein},
   booktitle=ECCV,
   year={2024}
}

@inproceedings{hong2024lrm,
   title={{LRM}: Large Reconstruction Model for Single Image to {3D}},
   author={Yicong Hong and Kai Zhang and Jiuxiang Gu and Sai Bi and Yang Zhou and Difan Liu and Feng Liu and Kalyan Sunkavalli and Trung Bui and Hao Tan},
   booktitle=ICLR,
   year={2024}
}

@inproceedings{tang2024lgm,
   title={{LGM}: Large Multi-View Gaussian Model for High-Resolution {3D} Content Creation},
   author={Jiaxiang Tang and Zhaoxi Chen and Xiaokang Chen and Tengfei Wang and Gang Zeng and Ziwei Liu},
   booktitle=ECCV,
   year={2024}
}

@inproceedings{xu2024dmv3d,
   title={{DMV3D}: Denoising Multi-View Diffusion using {3D} Large Reconstruction Model},
   author={Yinghao Xu and Hao Tan and Fujun Luan and Sai Bi and Peng Wang and Jiahao Li and Zifan Shi and Kalyan Sunkavalli and Gordon Wetzstein and Zexiang Xu and Kai Zhang},
   booktitle=ICLR,
   year={2024}
}

@inproceedings{li2024instant3d,
   title={Instant3D: Fast Text-to-{3D} with Sparse-View Generation and Large Reconstruction Model},
   author={Jiahao Li and Hao Tan and Kai Zhang and Zexiang Xu and Fujun Luan and Yinghao Xu and Yicong Hong and Kalyan Sunkavalli and Greg Shakhnarovich and Sai Bi},
   booktitle=ICLR,
   year={2024}
}

@inproceedings{wang2024pflrm,
   title={{PF-LRM}: Pose-Free Large Reconstruction Model for Joint Pose and Shape Prediction},
   author={Peng Wang and Hao Tan and Sai Bi and Yinghao Xu and Fujun Luan and Kalyan Sunkavalli and Wenping Wang and Zexiang Xu and Kai Zhang},
   booktitle=ICLR,
   year={2024}
}

@article{umeyama2002least,
  title={Least-squares estimation of transformation parameters between two point patterns},
  author={Umeyama, Shinji},
  journal=PAMI,
  volume={13},
  number={4},
  pages={376--380},
  year={2002},
  publisher={IEEE}
}

@inproceedings{xia2018gibson,
   title={Gibson Env: Real-World Perception for Embodied Agents},
   author={Fei Xia and Amir R. Zamir and Zhiyang He and Alexander Sax and Jitendra Malik and Silvio Savarese},
   booktitle=CVPR,
   year={2018}
}

@inproceedings{chang2017matterport3d,
   title={Matterport3D: Learning from {RGB-D} Data in Indoor Environments},
   author={Angel Chang and Angela Dai and Thomas Funkhouser and Maciej Halber and Matthias Nie{\ss}ner and Manolis Savva and Shuran Song and Andy Zeng and Yinda Zhang},
   booktitle={International Conference on 3D Vision (3DV)},
   year={2017}
}

@inproceedings{ramakrishnan2021hm3d,
   title={Habitat-Matterport {3D} Dataset ({HM3D}): 1000 Large-scale {3D} Environments for Embodied {AI}},
   author={Santhosh Kumar Ramakrishnan and Aaron Gokaslan and Erik Wijmans and Oleksandr Maksymets and Alexander Clegg and John Turner and Eric Undersander and Wojciech Galuba and Andrew Westbury and Angel X. Chang and Manolis Savva and Yili Zhao and Dhruv Batra},
   booktitle=NeurIPS,
   year={2021}
}

@inproceedings{savva2019habitat,
  title={Habitat: A platform for embodied ai research},
  author={Savva, Manolis and Kadian, Abhishek and Maksymets, Oleksandr and Zhao, Yili and Wijmans, Erik and Jain, Bhavana and Straub, Julian and Liu, Jia and Koltun, Vladlen and Malik, Jitendra and others},
  booktitle=ICCV,
  pages={9339--9347},
  year={2019}
}
}

\appendix
\renewcommand\thesection{\Alph{section}}
\setcounter{section}{0}
\renewcommand\thefigure{S\arabic{figure}}
\setcounter{figure}{0}
\renewcommand\thetable{S\arabic{table}}
\setcounter{table}{0}
\renewcommand\theequation{S\arabic{equation}}
\setcounter{equation}{0}

\section{Data Processing Pipeline}
\label{sec:data_processing}

Training a general-purpose streaming 3D reconstruction model requires
large-scale, diverse data with accurate ground-truth camera poses and
depth maps. Our training corpus (\cref{sec:training_data}) spans the
31 dataset entries listed in~\cref{tab:dataset_composition}. These
entries originate from heterogeneous sources with different data formats,
coordinate conventions, depth representations, and quality characteristics.
To unify them into a consistent training pipeline, we develop a multi-stage
data processing strategy:
(1)~\textit{Existing data processing} (\secref{sec:existing_data}):
standardizing publicly available datasets by converting
coordinate systems, normalizing depth scales, filtering corrupted
frames, and unifying metadata formats;
(2)~\textit{Synthetic data generation} (\secref{sec:gen_data}):
rendering additional training data from 3D environments with
controlled camera trajectories and pixel-perfect ground truth;
(3)~\textit{Gaming data processing} (\secref{sec:gaming_data}):
extracting high-quality pose and depth annotations from internal
game engine captures, which provide long trajectories through
large-scale, visually rich environments;
(4)~\textit{MatrixCity data sequencing} (\secref{sec:matrixcity_data}):
reorganizing the MatrixCity aerial and street-level data into
temporally continuous sequences via random walks on spatial
topologies, enabling its use for sequential training;
(5)~\textit{Cross-scene traversal data} (\secref{sec:traversal_data}):
rendering long-range cross-room RGBD video sequences from
large-scale 3D scene reconstructions using Habitat-Sim, providing
the multi-room traversal signals that existing indoor datasets lack.
Below we describe each pipeline in detail.

\subsection{Existing Data Processing}
\label{sec:existing_data}
To facilitate multi-dataset training, we standardize all publicly available datasets into a unified format through a series of preprocessing steps.

\textbf{Coordinate System Unification.} Datasets store camera poses using different
conventions.  We convert all poses to a consistent camera to world representation.
For datasets with non-standard axis conventions, such as UnReal4K~\cite{tosi2021smd},  an additional rotation matrix is applied to align the coordinate system with the OpenCV standard. Intrinsic parameters are either extracted from per-frame calibration files (\eg, .npz) or set to known constant values for datasets with a fixed camera model, such as TartanAir~\cite{wang2020tartanair}.

\textbf{Depth Scale Normalization.} The raw depth maps are provided in diverse formats and units. For instance, ScanNet and ScanNet++ store depth in millimeters as 16-bit PNG files, which are converted by dividing by 1000. VirtualKITTI2 uses centimeters stored in 16-bit color images, requiring division by 100. Other datasets, including DL3DV, HyperSim, TartanAir, and BlendedMVS, provide raw float .npy arrays in meters, while Waymo utilizes OpenEXR floating-point maps. All depth values are uniformly converted to meters and stored as float32.

\textbf{Corrupted Frame Filtering.} We apply several validation checks to remove corrupted or degenerate frames. First, scenes are discarded if the number of RGB images, depth maps, and camera pose files is inconsistent. Second, sequences with fewer than a minimum threshold of valid frames are excluded. Third, invalid depth values such as NaN or Inf are set to zero. Fourth, for datasets with noisy depth like DL3DV and MVS Synth, values exceeding the 98th percentile or an absolute threshold (\eg, 1000m) are clamped to zero. Fifth, for outdoor datasets such as DL3DV and Mapfree, sky regions are identified using pre-computed masks or depth saturation thresholds and their corresponding depth is set to zero. 
Finally, for DL3DV and Mapfree, pre-computed outlier masks are used to further remove geometrically inconsistent depth estimates. 

\textbf{Metadata Format Unification.} All datasets are converted into a common metadata structure, which is serialized as pickle files. This structure contains scene lists, per-frame index mappings (sceneids, id ranges), paths to images and depth maps, intrinsic matrices, and camera trajectories as 4$\times$4 pose matrices. This unified format is generated and cached during the initial execution, which enables efficient composition of multiple datasets at training time without requiring repeated parsing of the raw data.

\subsection{Generation Data Settings}
\label{sec:gen_data}

We render multi-view images of 3D assets in Objaverse~\cite{deitke2023objaverse} 
and Texverse~\cite{zhang2025texverse} using Blender Cycles. Each scene is normalized to fit within $[-0.9, 0.9]^3$. A pinhole camera with $\theta_h \sim \mathcal{U}(40^\circ, 70^\circ)$ and focal length $f = w / (2\tan(\theta_h/2))$ captures $512\times512$ images. Camera positions are sampled on a sphere around the origin at multiple elevations, all oriented toward the scene center. We apply HDR environment lighting and render RGBA color and metric depth in OpenEXR format. Images are tonemapped using percentile-based normalization with $\gamma{=}1/2.2$ correction.

\subsection{Gaming Data Processing Pipeline}
\label{sec:gaming_data}

\noindent We introduce a runtime acquisition pipeline that captures dense visual and geometric annotations from modern game engines. This provides a scalable source of large-scale, accurately annotated, and stylistically diverse long-sequence 3D data that is otherwise difficult to obtain from real-world captures alone. All sequences maintain high visual quality with smooth camera motion and sufficient inter-frame overlap, free of motion blur, over-exposure, or under-exposure artifacts. The dataset covers a wide variety of viewpoints including forward, lateral, and vertical gaze directions, and incorporates both intra-sequence and inter-sequence field-of-view variations to improve robustness to focal-length changes. Cutscenes, UI overlays, and other non-gameplay frames are excluded, and all graphics settings remain constant within each game to ensure consistent annotation quality across the entire dataset.

To ensure sufficient diversity in scene types and camera behaviors, we
establish a standardized collection protocol organized into two primary categories based on environment type:

\begin{itemize}
\item \textbf{Indoor scenes:} Covers navigation
  within enclosed environments such as buildings, rooms, and corridors.
  \begin{itemize}
  \item \textit{Free roaming:} Stochastic exploration along random
    trajectories, with frequent room transitions and floor changes;
  \item \textit{Loop roaming:} Departing from and returning to the same
    location to form closed-loop sequences through interconnected rooms;
  \item \textit{Transition navigation:} Traversing significant scene
    boundaries such as moving between distinct interiors or exiting to
    outdoor areas via doors and elevators.
  \end{itemize}

\item \textbf{Outdoor scenes:} Covers a broader
  range of navigation and observation modes in open environments.
  \begin{itemize}
  \item \textit{Free roaming:} Stochastic exploration along random
    trajectories, with diverse headings and frequent lateral viewpoint
    changes;
  \item \textit{Loop roaming:} Departing from and returning to the same
    location to form closed-loop sequences across open terrain;
  \item \textit{Transition navigation:} Traversing significant scene
    boundaries such as entering or exiting buildings and crossing between
    distinct outdoor regions;
  \item \textit{Dynamic sightseeing:} Touring scenes containing moving
    elements such as pedestrians, vehicles, or animals;
  \item \textit{Object orbiting:} Circling around objects to
    capture multi-view observations.
  \end{itemize}
\end{itemize}

\subsection{MatrixCity Data Sequencing}
\label{sec:matrixcity_data}

The MatrixCity dataset provides aerial and street-level multi-view images with known camera poses. However, the original data is organized for spatial coverage rather than temporal continuity---aerial images are sampled on a regular grid, while street images are stored as independent road segments. This is incompatible with downstream tasks that require temporally continuous input, such as video generation and sequential 3D reconstruction. We reorganize both data types into view-continuous sequences by performing random walks on their respective spatial topologies, without modifying the original images or poses.

\subsubsection{Aerial Data}
\label{sec:aerial_data}

\begin{figure}[t]
  \centering
  \begin{subfigure}[t]{0.48\linewidth}
    \centering
    \includegraphics[width=\linewidth]{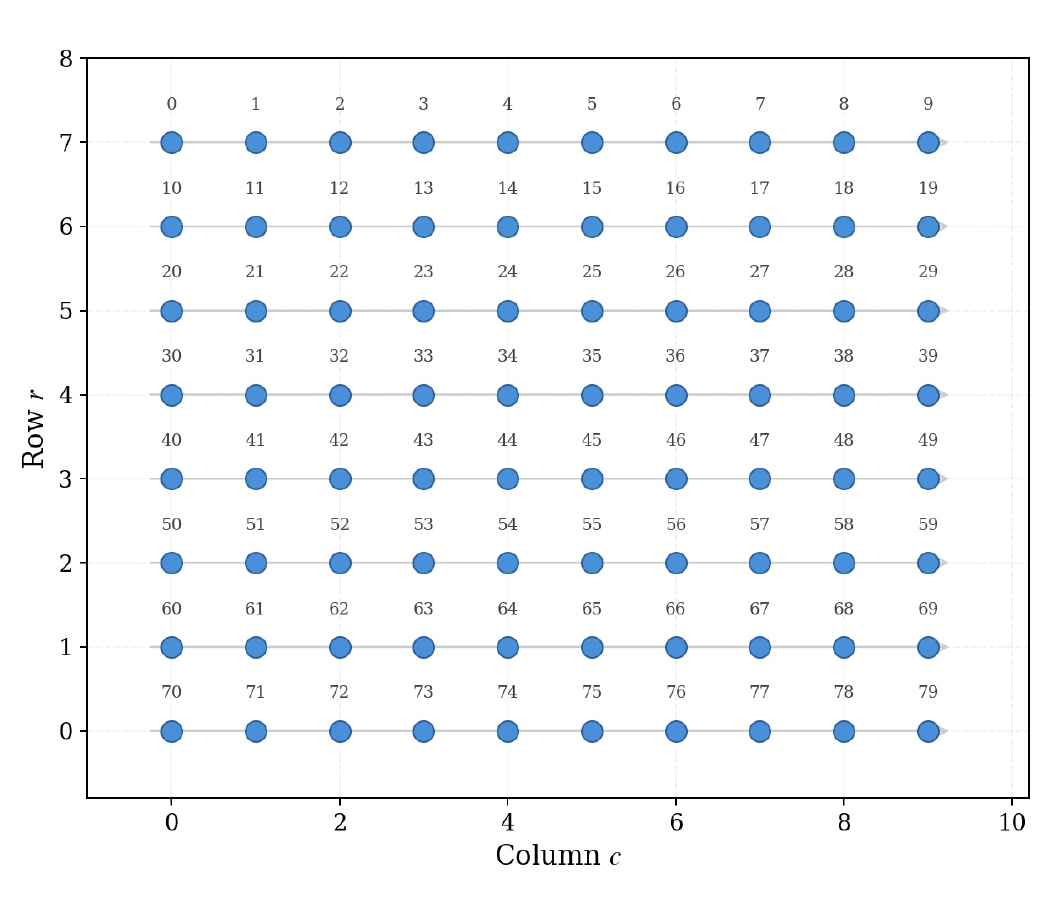}
    \caption{Row-major grid layout.}
    \label{fig:aerial_grid}
  \end{subfigure}
  \hfill
  \begin{subfigure}[t]{0.48\linewidth}
    \centering
    \includegraphics[width=\linewidth]{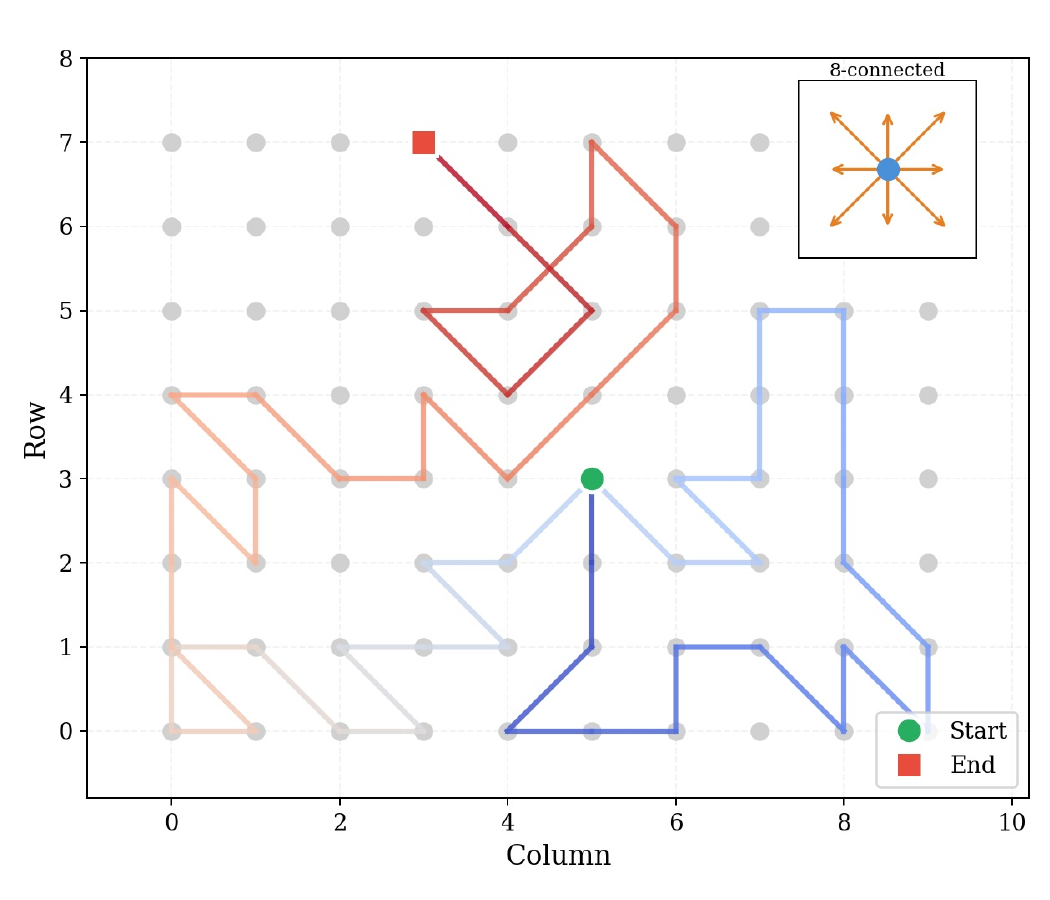}
    \caption{8-connected random walk.}
    \label{fig:aerial_walk}
  \end{subfigure}
  \caption{Aerial data sequencing. \textbf{(a)}~Each blue node is a camera position labeled by its row-major index ($\mathrm{index} = r \times W + c$); gray arrows show the storage scan order. \textbf{(b)}~A random walk on the grid; color gradient (blue $\to$ red) encodes temporal progression. The inset shows the 8-connected neighborhood. Each step excludes the previous position to suppress backtracking.}
  \label{fig:aerial}
\end{figure}

\paragraph{Data layout.}
Aerial data is arranged on an $H\!\times\!W$ regular grid in row-major order (\cref{fig:aerial}\,a). Each scene may contain multiple \emph{grid loops}, each corresponding to a complete grid sampling pass. The frame index~$i$ and grid coordinate $(r,c)$ satisfy
\begin{equation}
  r = \lfloor i / W \rfloor, \quad c = i \bmod W.
  \label{eq:grid_index}
\end{equation}
Consecutive frames are row-wise neighbors on the grid, but spatial discontinuities occur at row boundaries (\eg, the last frame of one row and the first frame of the next row are far apart), so the raw storage order does not form a natural camera trajectory.

\paragraph{Sequencing method.}
We treat the grid as a graph with 8-connected adjacency and perform random walks to produce spatially continuous trajectories (Fig.~\ref{fig:aerial}\,b). To suppress immediate backtracking, each step excludes the previous position from the candidate set. Grid coordinates are mapped to global frame indices via
\begin{equation}
  \mathrm{idx} = l \cdot N_{\mathrm{loop}} + r \cdot W + c,
  \label{eq:global_index}
\end{equation}
where $l$ is the grid loop index and $N_{\mathrm{loop}}$ is the number of frames per loop. The number of sequences is proportional to the data size (by default, 5 sequences per 1{,}000 frames). The full procedure is given in Algorithm~\ref{alg:aerial}.

\begin{algorithm}[t]
\caption{Aerial Grid Random Walk}
\label{alg:aerial}
\begin{algorithmic}[1]
\Require Grid size $H\!\times\!W$, sequence length $T$, loop index $l$
\Ensure Frame index sequence $\mathcal{S}$
\State Sample random start position $(r, c) \in [0,H) \times [0,W)$
\State $\mathrm{prev} \gets \texttt{null}$
\For{$t = 1$ \textbf{to} $T$}
    \State $\mathcal{S}.\text{append}(l \cdot N_{\mathrm{loop}} + r \cdot W + c)$
    \State $\mathcal{N} \gets \{(r\!+\!\Delta r,\, c\!+\!\Delta c) \mid (\Delta r, \Delta c) \in \mathcal{D}_8\} \cap \mathcal{G} \;\setminus\; \{\mathrm{prev}\}$
    \If{$\mathcal{N} = \emptyset$}
        \State $\mathcal{N} \gets$ all valid 8-connected neighbors \Comment{corner fallback}
    \EndIf
    \State $\mathrm{prev} \gets (r, c)$
    \State $(r, c) \gets \textsc{RandomChoice}(\mathcal{N})$
\EndFor
\State \Return $\mathcal{S}$
\end{algorithmic}
\end{algorithm}

\subsubsection{Street Data}
\label{sec:street_data}

\paragraph{Data layout.}
Street data is captured per-street. Each street is a linear trajectory with $N$ evenly-spaced camera positions, each photographed from 5 viewpoints (viewpoints 0--3 are horizontal; viewpoint~4 is zenith-facing), yielding $5N$ frames per street. Frames are stored in viewpoint-first order: all $N$ positions of viewpoint~0 appear first, followed by viewpoints 1--4 in sequence. Different streets may physically intersect (\eg, at crossroads), but are stored independently without explicit connectivity information.

\begin{figure}[t]
  \centering
  \begin{subfigure}[t]{0.32\linewidth}
    \centering
    \includegraphics[width=\linewidth]{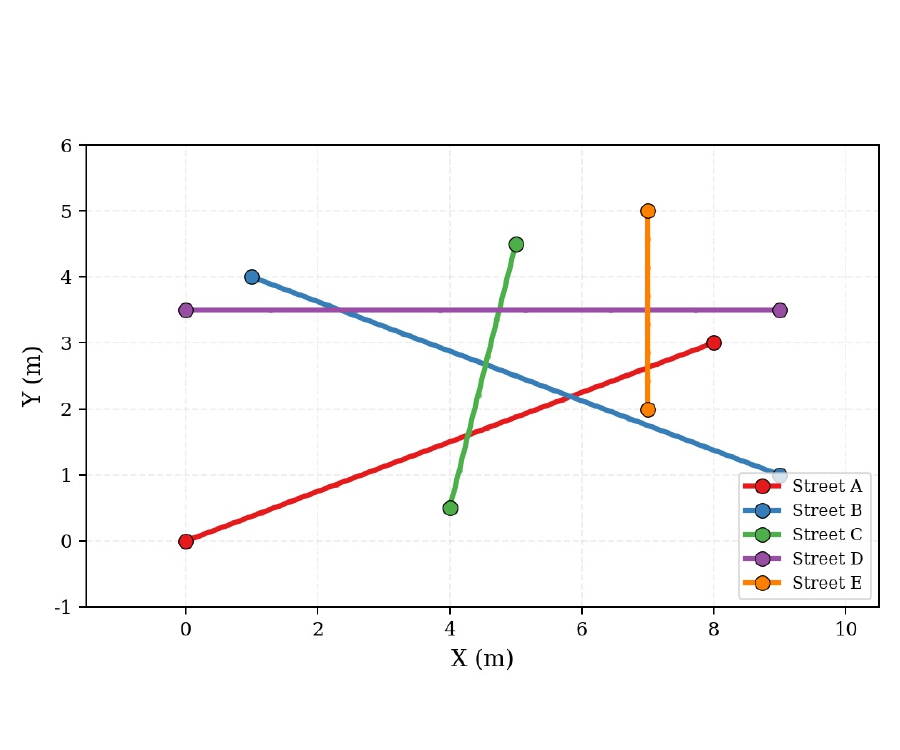}
    \caption{Street identification.}
    \label{fig:street_stage1}
  \end{subfigure}
  \hfill
  \begin{subfigure}[t]{0.32\linewidth}
    \centering
    \includegraphics[width=\linewidth]{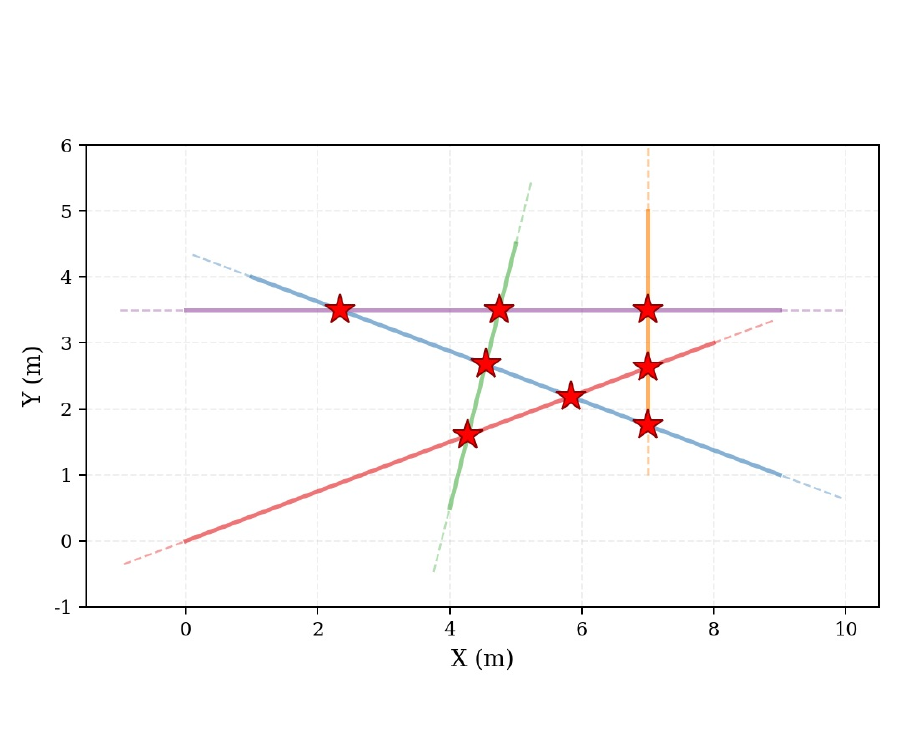}
    \caption{Intersection detection.}
    \label{fig:street_stage2}
  \end{subfigure}
  \hfill
  \begin{subfigure}[t]{0.32\linewidth}
    \centering
    \includegraphics[width=\linewidth]{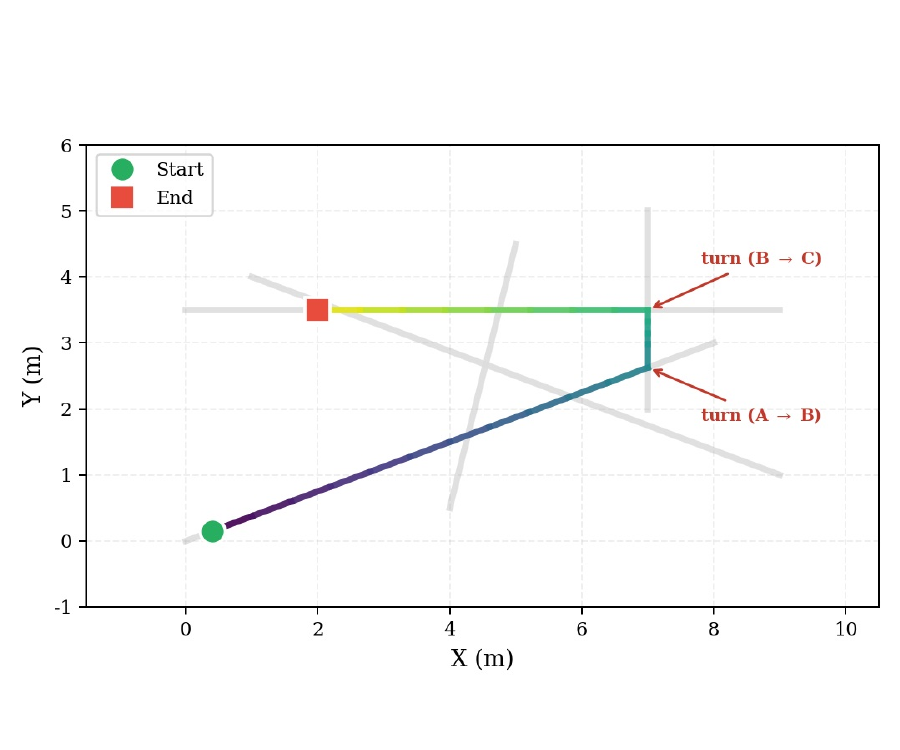}
    \caption{Graph-based random walk.}
    \label{fig:street_stage3}
  \end{subfigure}
  \caption{Street data sequencing pipeline. \textbf{(a)}~Stage~1: independently identified street segments (A--E) with camera positions (dots). \textbf{(b)}~Stage~2: red stars mark detected intersections; dashed lines show optional segment extensions for near-miss junctions. \textbf{(c)}~Stage~3: a random walk traversing Street~A $\to$ B $\to$ C via intersections; color gradient encodes temporal order.}
  \label{fig:street}
\end{figure}

Sequencing proceeds in three stages (\cref{fig:street}): street identification, intersection detection, and graph-based random walk.

\paragraph{Stage 1: Street identification.}
We exploit the viewpoint-first storage order to automatically segment street boundaries via \emph{position repetition detection} (\cref{fig:street}\,a). Frames are traversed sequentially, and camera positions are extracted. A set of unique positions $\mathcal{U}$ is maintained; when a new position falls within $\epsilon = 0.01$\,m of any existing point in $\mathcal{U}$, it is flagged as a duplicate, indicating that viewpoint~0 traversal is complete. At this point $|\mathcal{U}| = N$ gives the number of unique positions for that street, and the subsequent $5N$ frames are partitioned into 5 viewpoint segments.

\paragraph{Stage 2: Intersection detection and graph construction.}
To establish topological connectivity, each street is simplified to its endpoint-to-endpoint 2D line segment (projected onto the $XY$ plane), and all segment pairs are tested for intersection (\cref{fig:street}\,b). To handle streets that are physically close but do not precisely intersect (\eg, T-junctions), each segment is optionally extended by a configurable distance $d_{\mathrm{ext}}$ at both ends before testing. Detected intersections define an undirected street connectivity graph $G = (V, E)$, where vertices~$V$ are streets and edges~$E$ connect intersecting pairs.

\paragraph{Stage 3: Graph-based random walk.}
We generate cross-street continuous sequences by performing random walks on $G$ (\cref{alg:street},~\cref{fig:street}\,c). Three key design choices ensure sequence quality:
\begin{enumerate}[leftmargin=*,nosep]
  \item \textbf{Intersection decisions.} When the walker approaches an intersection (distance $< d_{\mathrm{prox}}$), it switches to a connecting street with probability $p_{\mathrm{turn}} = 0.5$.
  \item \textbf{Viewpoint continuity.} Upon switching streets, the viewpoint whose camera forward direction (third column of the rotation matrix) best matches the current one is selected:
  \begin{equation}
    v^* = \arg\max_{v \in \{0,\dots,4\}} \; \mathbf{f}_{\mathrm{curr}} \cdot \mathbf{f}_{v}.
    \label{eq:viewpoint_select}
  \end{equation}
  \item \textbf{Endpoint handling.} At dead ends (degree $\leq 1$), the walking direction is reversed.
\end{enumerate}

\begin{algorithm}[t]
\caption{Street Network Random Walk}
\label{alg:street}
\begin{algorithmic}[1]
\Require Street graph $G$, sequence length $T$, turn probability $p_{\mathrm{turn}}$, proximity threshold $d_{\mathrm{prox}}$
\Ensure Frame sequence $\mathcal{S}$
\State Randomly select start street $e$, position index $i$, viewpoint $v \in \{0..3\}$, direction $d \in \{+1, -1\}$
\For{$t = 1$ \textbf{to} $T$}
    \State $\mathcal{S}.\text{append}\bigl(\textsc{Frame}(e, i, v)\bigr)$
    \State $i_{\mathrm{next}} \gets i + d$
    \If{$i_{\mathrm{next}}$ out of bounds} \Comment{street endpoint}
        \If{$\deg(e) \leq 1$}
            \State $d \gets -d$ \Comment{dead end: reverse}
        \Else
            \State $(e, i, v, d) \gets \textsc{SwitchStreet}(e, i, v)$
        \EndIf
    \ElsIf{$\exists$ intersection within $d_{\mathrm{prox}}$ \textbf{and} $\textsc{Rand}() < p_{\mathrm{turn}}$}
        \State $(e, i, v, d) \gets \textsc{SwitchStreet}(e, i, v)$
    \Else
        \State $i \gets i_{\mathrm{next}}$ \Comment{advance along current street}
    \EndIf
\EndFor
\State \Return $\mathcal{S}$
\Statex
\Function{SwitchStreet}{$e, i, v$}
    \State $e' \gets$ random neighbor of $e$ in $G$
    \State $i' \gets$ nearest position on $e'$ to intersection point
    \State $v' \gets \arg\max_{v''} \mathbf{f}_{\mathrm{curr}} \cdot \mathbf{f}_{v''}$ \Comment{viewpoint continuity}
    \State $d' \gets$ direction away from intersection
    \State \Return $(e', i', v', d')$
\EndFunction
\end{algorithmic}
\end{algorithm}

\subsection{Cross-Scene Traversal Data Construction}
\label{sec:traversal_data}

Long-range streaming 3D reconstruction requires the model to handle continuously expanding observation sequences in which the camera moves from one room through a corridor into the next, encountering substantial changes in geometry and appearance along the way. Existing indoor RGBD datasets (\eg, ScanNet~\cite{dai2017scannet}) are largely confined to single rooms or small regions and therefore lack the cross-region, long-range traversal signals needed for training. To address this limitation, we render cross-room continuous RGBD video sequences from large-scale real-world 3D scene reconstructions using Habitat-Sim~\cite{savva2019habitat}.

\paragraph{Scene sources.}
We draw scenes from three complementary indoor 3D datasets: Gibson~\cite{xia2018gibson} (${\sim}$450 building-scale scans covering residential and office environments with high geometric fidelity), Matterport3D~\cite{chang2017matterport3d} (90 large residential and commercial spaces, each typically containing 10 to 30 rooms with rich furniture layouts and multi-story structures), and HM3D~\cite{ramakrishnan2021hm3d} (900 semantically diverse indoor environments spanning a wide range of scene scales and complexities). The three datasets differ in capture devices, reconstruction quality, and scene types, and their joint use broadens the training distribution while mitigating overfitting to any single data source.

\paragraph{Trajectory generation.}
For each scene, we first compute a navigation mesh and then randomly sample $N{+}1$ waypoints on walkable surfaces, requiring a minimum geodesic distance of 2\,m between consecutive waypoints to ensure sufficient spatial coverage. The number of segments $N$ is drawn from a piecewise mixture distribution (40\% in 5 to 8, 45\% in 8 to 12, and 15\% in 12 to 16), corresponding to local exploration, cross-functional-area walking, and full-scene traversal, respectively. Adjacent waypoints are connected via the shortest geodesic path on the navigation mesh, allowing trajectories to naturally pass through doorways and corridors for cross-room traversal. The concatenated path is then linearly interpolated, smoothed with a double-pass moving average, and projected back onto the navigation mesh, yielding a dense, smooth, and physically plausible trajectory. A single trajectory typically comprises 500 to 3{,}000 frames (roughly 15 to 100\,s at 30\,fps).

\paragraph{Motion control.}
To produce realistic egocentric videos, we design a multi-stage motion controller that converts the trajectory path into per-frame camera poses. Translation is governed by a first-order low-pass filter ($\alpha{=}0.18$) combined with navigation-mesh collision correction. Yaw is steered toward a lookahead target 10 frames ahead through a $\tanh$-saturated velocity model that provides natural acceleration and deceleration during turns. Pitch follows terrain elevation changes and is smoothed by exponential decay. On top of this base motion, IIR-filtered Gaussian noise is added to both position and orientation to simulate handheld camera shake, and stochastic glance events, \ie brief lateral and vertical head turns following a half-cosine envelope, are triggered at random to reproduce the natural gaze shifts that occur during walking.

\paragraph{Parameter diversification.}
Camera intrinsics and motion parameters are independently sampled per sequence from piecewise mixture uniform distributions to cover a broad range of real-world conditions: horizontal field of view from 40\textdegree{} to 100\textdegree, sensor height from 1.35 to 1.80\,m, walking speed from 0.3 to 1.8\,m/s, and several tiers of jitter intensity and glance frequency. Walking speed is sampled independently per trajectory segment, producing natural speed variations within a single sequence. All randomized parameters and waypoint coordinates are recorded in a configuration file for exact reproducibility.

\begin{figure*}[t]
  \centering
  \setlength{\tabcolsep}{1.5pt}
  \adjustbox{max width=\linewidth}{%
  \begin{tabular}{ccc}
    \includegraphics[height=0.2\linewidth]{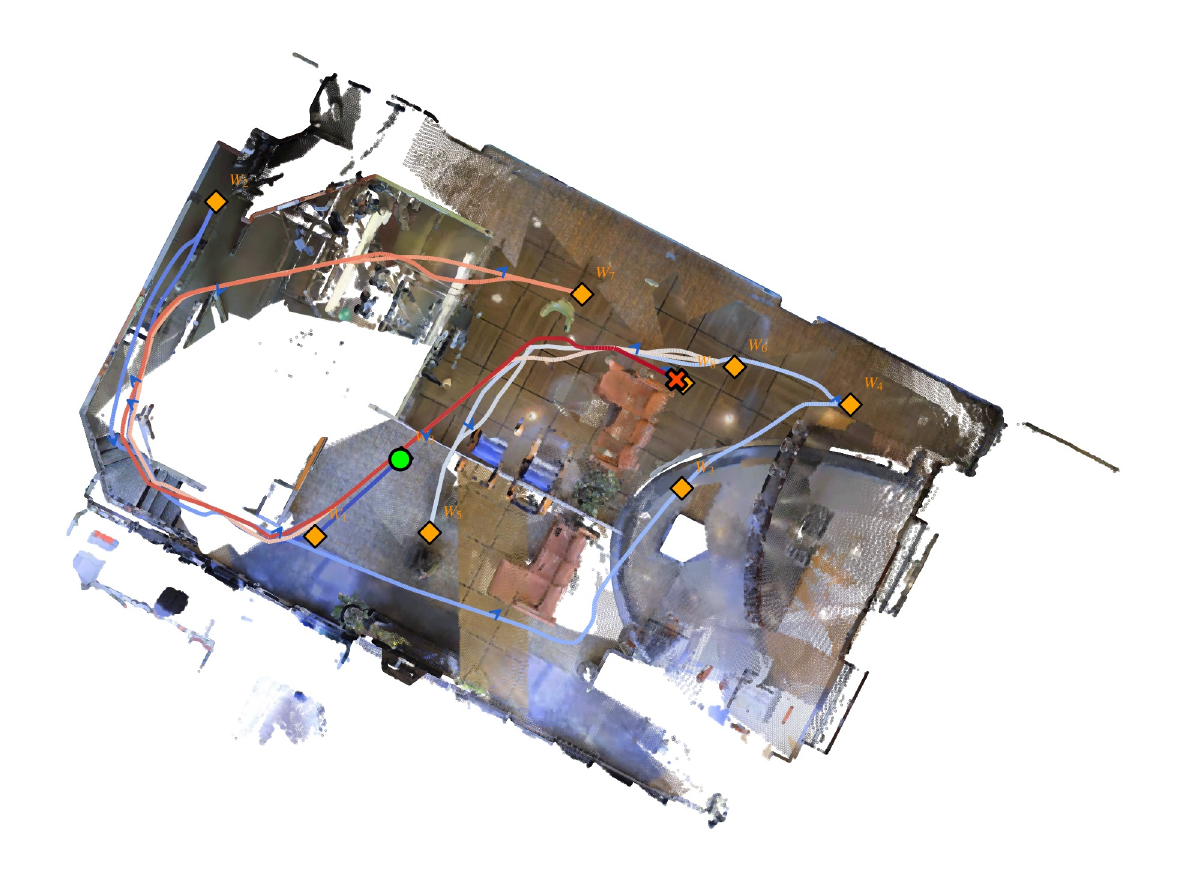} &
    \includegraphics[height=0.2\linewidth]{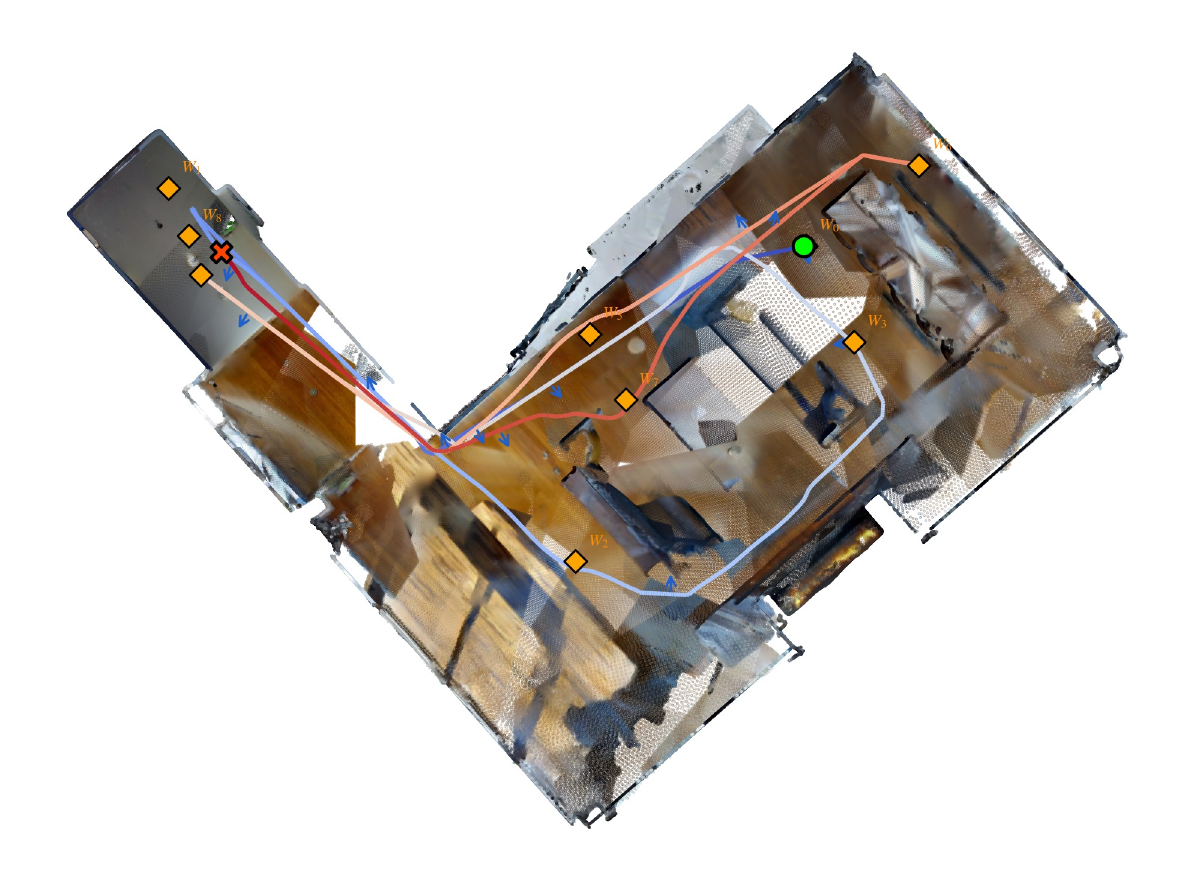} &
    \includegraphics[height=0.2\linewidth]{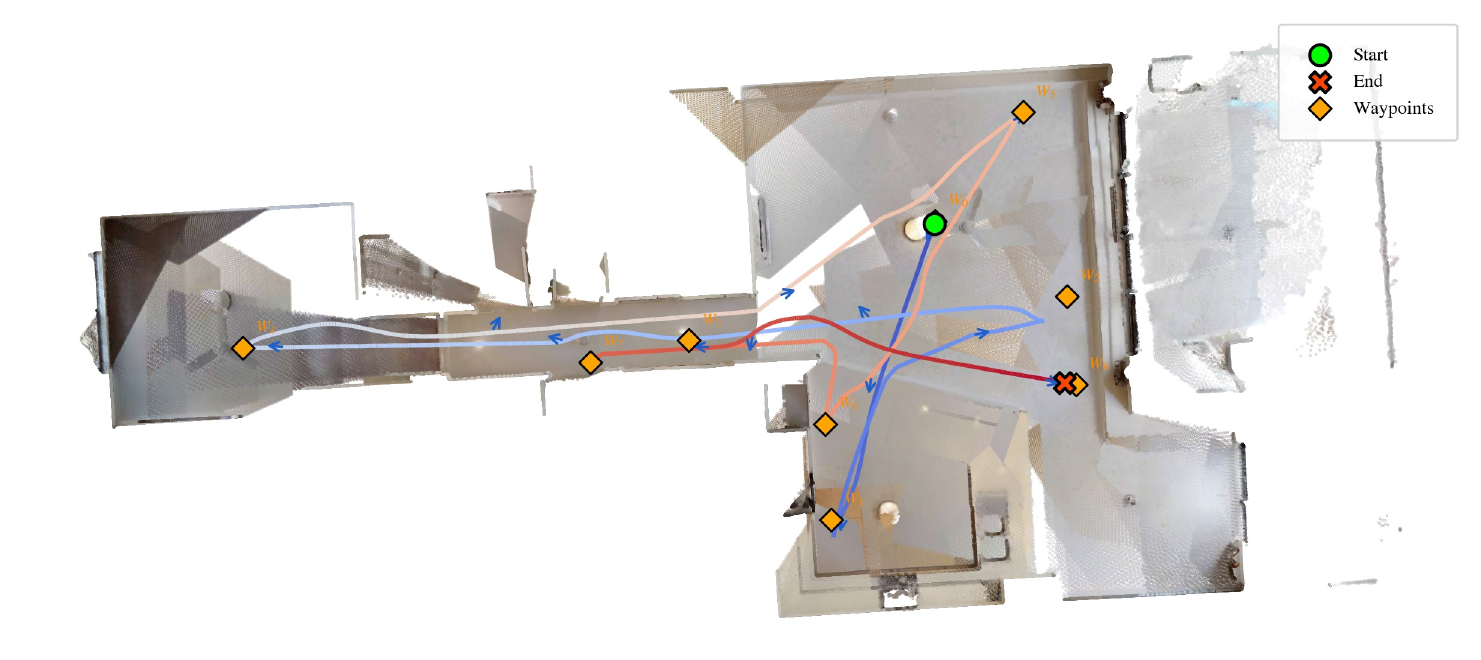} \\[2pt]
    \includegraphics[height=0.2\linewidth]{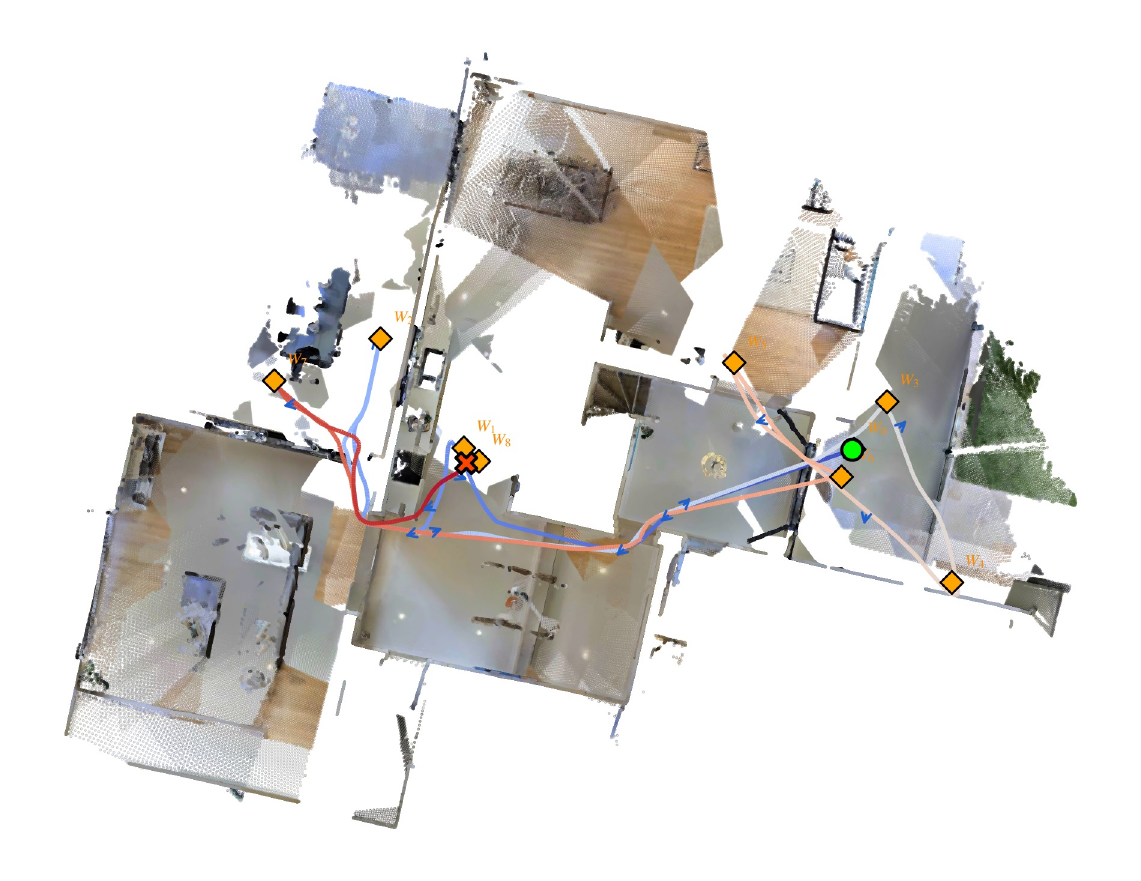} &
    \includegraphics[height=0.2\linewidth]{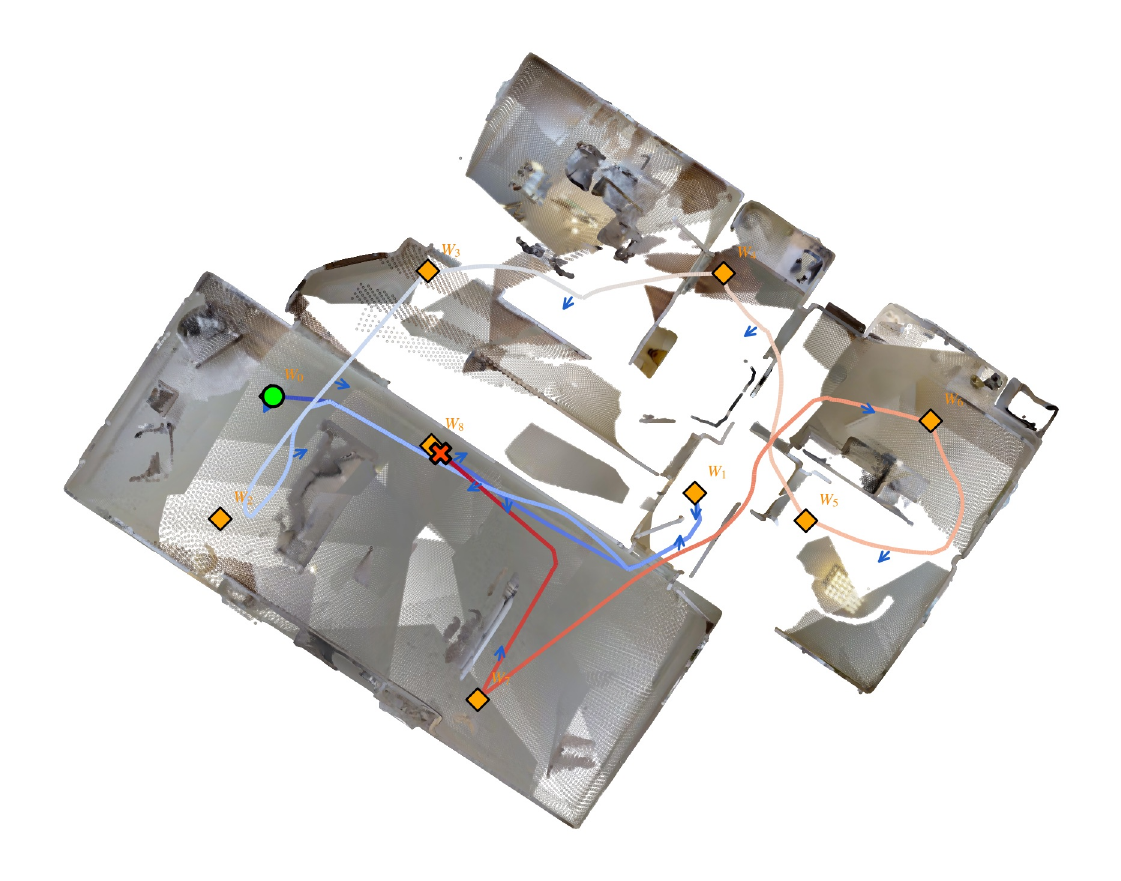} &
    \includegraphics[height=0.2\linewidth]{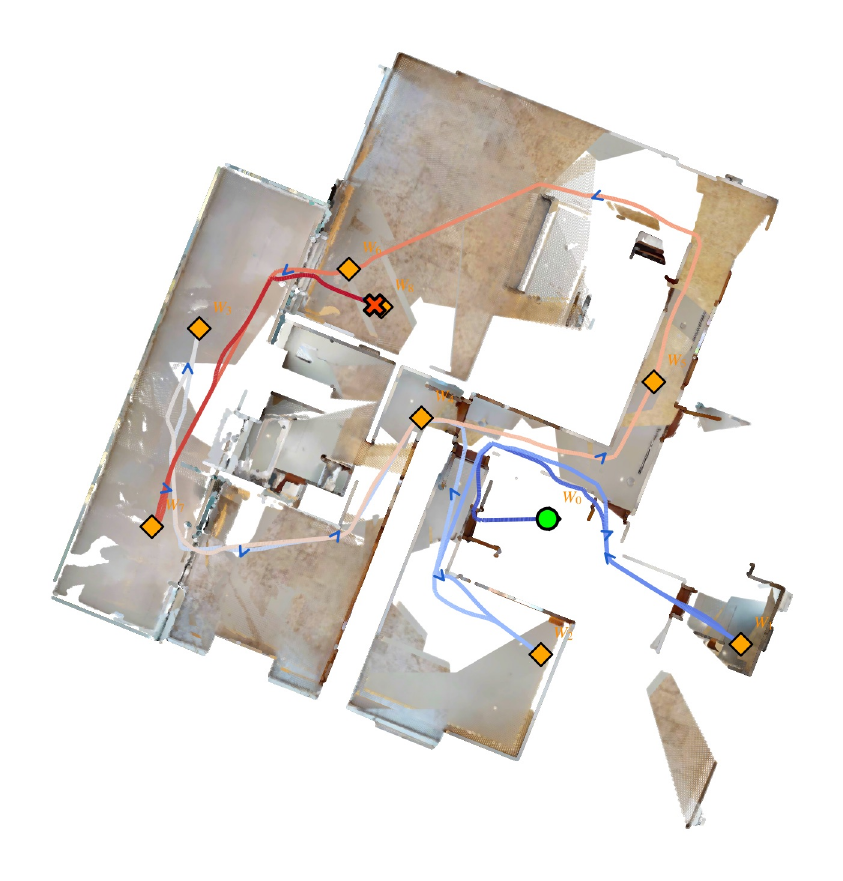} \\[2pt]
    {\small Matterport3D} & {\small Gibson} & {\small HM3D} \\
  \end{tabular}}
  \caption{%
    Top-down visualizations of rendered point clouds and sampled camera trajectories across six scenes from three datasets (Matterport3D, Gibson, and HM3D).
    Each point cloud is colored using per-pixel RGB values from the rendered frames.
    The trajectory color varies from \textcolor{blue}{blue} (start) to \textcolor{red}{red} (end), indicating temporal progression.
    \textcolor{green}{Green} circles and \textcolor{red}{red} crosses mark the start and end positions, respectively.
    \textcolor{orange}{Orange} diamonds denote randomly sampled waypoints on the navigation mesh, and \textcolor{blue}{blue} arrows indicate the camera viewing direction at regular intervals.
  }
  \label{fig:traversal_viz}
\end{figure*}

\paragraph{Rendering and output.}
We employ a pinhole camera model and render at $640{\times}480$ resolution and 30\,fps, producing per-frame RGB images and metric depth maps together with 6-DoF camera poses (position and quaternion) and intrinsics ($f_x, f_y, c_x, c_y$). In total, we generate approximately 2{,}800 sequences from the three datasets, each containing 1k to 5k frames, amounting to 14.4\,TB of data. \Cref{fig:traversal_viz} visualizes the top-down point clouds and sampled trajectories for several representative scenes.

\end{document}